\documentclass{article}

    \PassOptionsToPackage{numbers, compress}{natbib}

 \usepackage[main, final]{neurips_2026}

\usepackage[utf8]{inputenc}
\usepackage[T1]{fontenc}
\usepackage{hyperref}
\usepackage{url}
\usepackage{booktabs}
\usepackage{amsfonts}
\usepackage{nicefrac}
\usepackage{microtype}
\usepackage{xcolor}
\usepackage{graphicx}
\usepackage{subcaption}
\usepackage{amsmath}
\usepackage{multirow}
\usepackage{enumitem}
\usepackage{xspace}

\newcommand{\method}{\textsc{C$^3$L}\xspace}

\title{Semantic Concept Steering Breaks the Explanation Drift Loop in Continual Learning}

\author{%
  Yehonatan Elisha \\
  Tel Aviv University \\
  \AND
  Oren Barkan \\
  The Open University \\
  \And
  Noam Koenigstein \\
  Tel Aviv University \\
}

\begin{document}

\maketitle

\begin{abstract}
Catastrophic forgetting in continual learning (CL) manifests not only as accuracy degradation but also as \emph{explanation drift}: models maintain predictive accuracy while silently shifting attention from diagnostic features to spurious correlations. This work identifies and resolves a previously overlooked structural flaw in explanation-aware CL: pixel-level saliency regularization derives supervision targets from the model's own frozen saliency maps, creating a \emph{self-referential drift loop} that actively reinforces spurious attention across tasks. We introduce \textbf{C$^3$L} (Concept-Consistent Continual Learning), which breaks this loop by anchoring regularization to class-consistent semantic concepts
(e.g., ``curved beak'', ``striped pattern'') discovered offline, independent of model state, providing (i) target independence from model drift, and (ii) cross-instance
semantic consistency across every instance of a class. We identify that our concept-based objective, combined with approximate LRP relevance conservation, induces a \emph{crowding-out effect}: since concept and background regions compete for the relevance of the same prediction, enforcing high relevance within concept regions implicitly suppresses spurious background correlations without explicit penalization. \method does not require human-annotated concept sets, as it employs a fully automated concept discovery and spatial grounding. Extensive evaluation across five benchmarks and over 20 baselines demonstrates state-of-the-art performance in accuracy, forgetting, explanation quality, spurious correlation mitigation, and out-of-distribution robustness.
Our code is provided at: \url{https://github.com/yonisGit/concept_consistent_continual_learning}
\end{abstract}

\vspace{-3mm}

\begin{figure}[h!]
  \centering
  \includegraphics[width=0.86\textwidth]{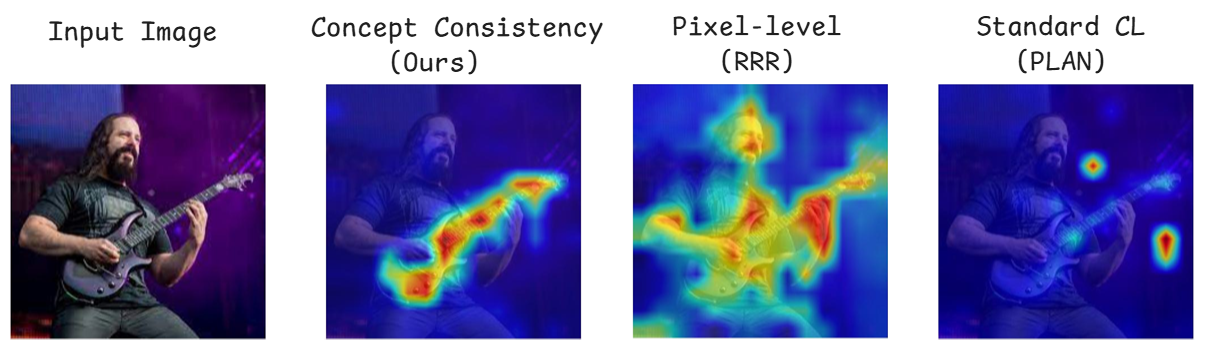}
  \caption{\textbf{Mitigating Explanation Drift via Semantic Anchoring.} Saliency maps for ``Electric Guitar'' sample after continual learning. Standard CL (PLAN) suffers from explanation drift, shifting attention to spurious background artifacts. While pixel-level methods (RRR) attempt to mitigate this, they result in noisy, non-specific attributions. Our \method anchors model representations to semantically meaningful concepts, maintaining focus on diagnostic features like the fretboard and guitar body.}
  \label{fig:method_overview}
\end{figure}
\vspace{-2mm}

\section{Introduction}

Deep neural networks trained on static datasets catastrophically forget previously learned knowledge when adapted to sequential tasks~\citep{mccloskey1989catastrophic,french1999catastrophic}. This phenomenon poses fundamental challenges for deploying adaptive systems in dynamic environments, from medical diagnosis systems learning new diseases to autonomous vehicles adapting to novel scenarios~\citep{parisi2019continual}. While the continual learning literature has developed sophisticated mechanisms to preserve predictive accuracy through replay~\citep{rebuffi2017icarl,buzzega2020dark}, regularization~\citep{kirkpatrick2017overcoming,zenke2017continual}, and architectural adaptations~\citep{rusu2016progressive,wang2022learning}, these approaches treat models as black boxes operating in parameter or feature space - preserving \emph{what} is predicted, not \emph{how}.

\textbf{The explanation drift problem.}
\citet{ebrahimi2020remembering} showed that catastrophic forgetting manifests not only as accuracy degradation but also as \emph{explanation drift}: models maintain accuracy while
silently shifting attention from diagnostic features (e.g., bird beak shapes) to spurious correlations (e.g., background foliage). This instability in reasoning is not captured by standard metrics, yet has critical implications - when such spurious correlations fail under distribution shift, performance can collapse~\citep{geirhos2018imagenet,Elisha_2026_CVPR}. Existing explanation-aware CL methods address this drift by regularizing pixel-level saliency maps~\citep{ebrahimi2020remembering,saha2023saliency,liu2024task}, but they share a fundamental and previously overlooked structural flaw: Pixel-level methods derive supervision targets from the model's own frozen saliency at buffer insertion. If the model has already drifted toward spurious features at that point, these patterns become the regularization target, effectively reinforcing the drift across future tasks.
This \emph{self-referential drift loop} is not merely an attribution noise issue - it is a structural property of any scheme that derives targets from model-state-dependent saliency.

\begin{figure}[h]
  \centering
  \includegraphics[width=0.980\textwidth]{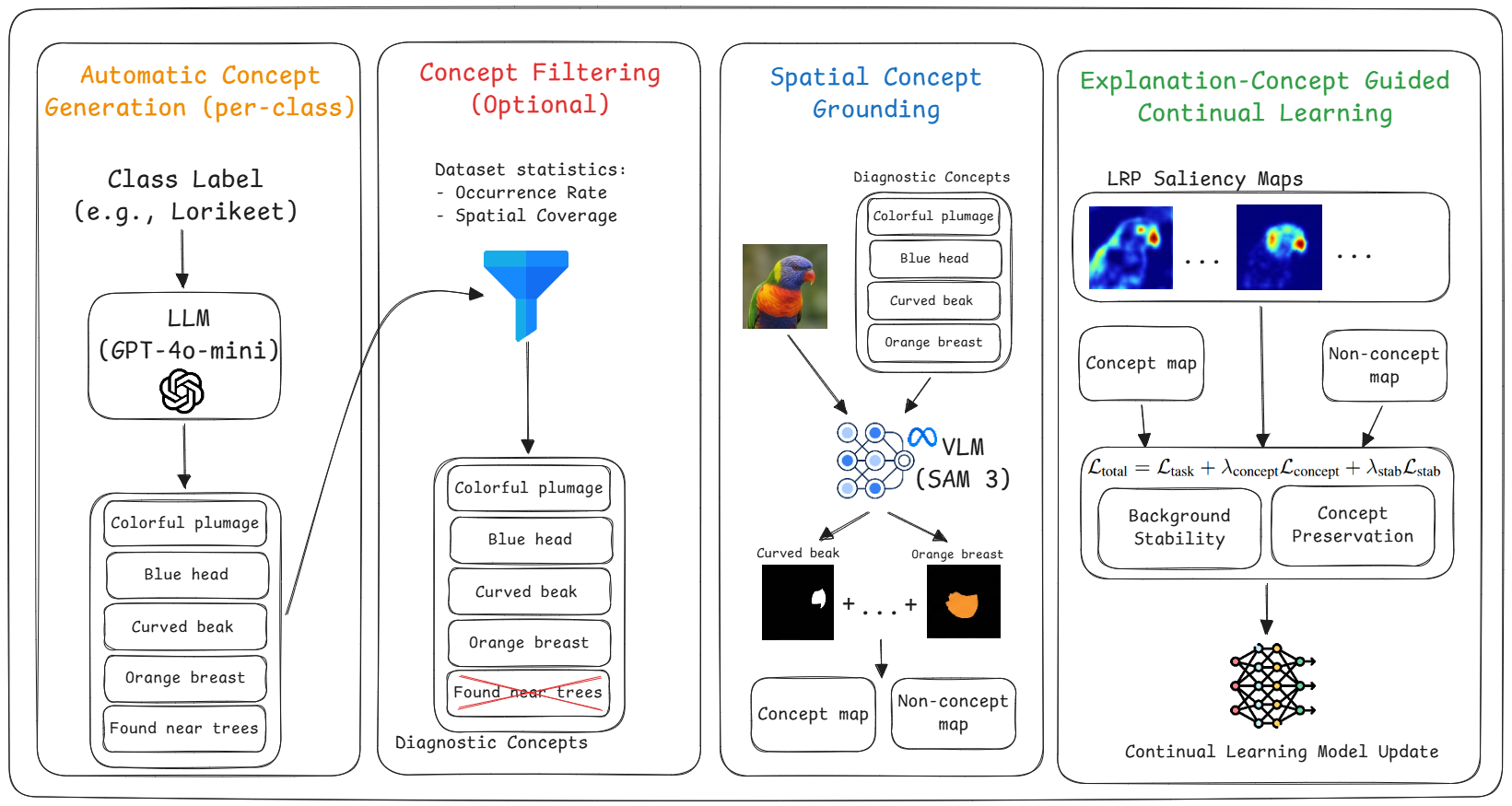}
  \caption{\textbf{\method Overview.}
Discriminative visual concepts are discovered offline via an LLM and spatially
grounded using SAM~3, producing model-state-independent supervision targets that
break the self-referential drift loop of prior pixel-level methods. During continual
training, $\mathcal{L}_{\text{concept}}$ anchors relevance to semantic regions via
a \emph{crowding-out effect}, while $\mathcal{L}_{\text{stab}}$ preserves
background spatial structure against task-specific drift.}
  \label{fig:method_diagram}
\end{figure}

\textbf{Breaking the loop with semantic concept anchoring.}
We present \textbf{C$^3$L} (Concept-Consistent Continual Learning) (Figure~\ref{fig:method_diagram}), which breaks this loop by grounding regularization in \emph{class-consistent semantic concepts} - human-interpretable visual attributes (e.g., ``curved beak'') discovered \emph{offline}, prior to and independent of any model state. This yields two structural
properties no previous explanation-aware method can provide: \emph{(i) target independence} - concept targets cannot encode model drift as they are produced independently of the learner's state; and \emph{(ii) cross-instance semantic consistency} - every instance of a class is regularized toward the same diagnostic attributes, inducing a class-level invariant beyond instance-specific saliency supervision.

Furthermore, we identify a \emph{crowding-out effect}: under approximate LRP relevance conservation, concept and background regions compete for the relevance of the same prediction. Enforcing high relevance within semantically grounded concept regions therefore implicitly suppresses spurious background attributions, without any explicit penalization, turning semantic anchoring into an implicit suppression mechanism, which we validate empirically. A complementary background stability term preserves the \emph{spatial structure} of background relevance while allowing
its magnitude to shrink, preventing freed relevance mass from drifting toward new task-specific artifacts rather than toward diagnostic concepts. Finally, a fully automated
framework, leveraging LLMs for concept discovery~\citep{oikarinen2023label} and open-vocabulary segmentation for spatial grounding~\citep{carion2025sam}, eliminates the manual annotation burden that has historically prevented concept-based approaches~\citep{koh2020concept} from scaling to the many-class, sequential-task regime.

\textbf{Contributions:}
\begin{enumerate}[nosep]
    \item \textbf{Problem identification:} We formalize the self-referential drift loop as a structural obstacle in pixel-level explanation-aware CL (Section~\ref{sec:method}).

    \item \textbf{Crowding-out mechanism:} We identify the crowding-out effect induced by approximate LRP conservation, which turns semantic anchoring into an implicit background-suppression mechanism, and validate it quantitatively.

    \item \textbf{C$^3$L framework:} A novel concept-guided continual learning framework. It operates without manual annotation and leverages semantic anchors to eliminate self-referential drift and drive the generalization of transferable concepts in the model (\S\ref{subsec:robust}).

    \item \textbf{State-of-the-art results:} C$^3$L outperforms nearly twenty baselines across five benchmarks on accuracy, forgetting, semantic localization, spurious correlation mitigation, and OOD robustness, on both CNN and ViT architectures.
\end{enumerate}
\vspace{-3mm}

\section{Related Work}

\textbf{Continual learning strategies.}
CL methods can be roughly divided into several types. Replay-based methods \cite{rebuffi2017icarl,buzzega2020dark,caccia2022new,park2024coreset} store exemplars and interleave them during new task training. Wan et al.~\citep{wan2025understanding} show that replay is essential for maintaining feature learning dynamics across tasks. Regularization-based methods \cite{kirkpatrick2017overcoming,zenke2017continual,kim2024prediction} add parameter regularization terms or function regularization terms to transfer knowledge from the old model to the new one. Architecture-based methods often add new trainable modules to expand the feature space for new tasks. Prompt-based \cite{wang2022learning,wang2022dualprompt,smith2023coda} methods keep the backbone frozen and learn task-specific input prompts. Adapter-based methods~\citep{chen2025mos,wang2025min,zhao2025plan} insert learnable modules or selectively fine-tune components while freezing pretrained weights. All these operate in parameter, feature, or sample space without preserving \emph{semantic reasoning} - which visual features the model attends to. This leaves models vulnerable to explanation drift \cite{ebrahimi2020remembering}.

\textbf{Explainable AI and saliency maps.}
Explainability methods aim to reveal which input features drive a model's prediction. Model-agnostic approaches such as LIME~\citep{ribeiro2016lime} and SHAP~\citep{lundberg2017shap}, as well as learned explainers~\citep{barkan2023ltx}, apply across domains, and dedicated attribution techniques have been developed for recommender systems~\citep{barkan2024counterfactual,gurevitch2025lxr,baklanov2025refining,barkan2026fidelity,arviv2026monosemantic} and for language models~\citep{barkan2021gradsam,barkan2024learning,barkan2024amask,barkan2024rpi}. In computer vision, explanations are typically given as saliency maps, produced by gradient-based methods such as Grad-CAM~\citep{selvaraju2017grad} and its extensions to similarity models~\citep{barkan2021gam} and transformers~\citep{chefer2021transformer}; by relevance-propagation methods that approximately conserve relevance, such as LRP~\citep{bach2015pixel,montavon2019layer} and AttnLRP~\citep{achtibat2024attnlrp}; and by path-integration methods originating from Integrated Gradients~\citep{sundararajan2017axiomatic}, later extended to integration over intermediate representations~\citep{barkan2023visual,barkan2023dix}, stochastic and probabilistic paths and baselines~\citep{barkan2023six,elisha2024ppi}, and metric-adapted baselines~\citep{barkan2025bee}. Related work revisits the completeness property underlying conservation-based attribution~\citep{weisshaddad2025slc} and the human-aligned evaluation of saliency maps~\citep{elisha2026rethinking}. Concept-based explanations link attributions to human-interpretable concepts~\citep{kim2018interpretability,achtibat2023crp,elisha2026conex}, and concept-guided relevance has recently been used as a \emph{training} signal to steer models away from spurious correlations in a single fine-tuning stage, by explicitly suppressing background relevance~\citep{Elisha_2026_CVPR}. \textbf{C$^3$L} differs in both setting and mechanism: it targets continual learning, where explanations drift as the model is updated on a sequence of tasks; it identifies the self-referential drift loop of pixel-level explanation-aware CL methods and replaces their model-derived targets with learner-independent concept targets; and rather than suppressing the background, it stabilizes its spatial structure with a scale-invariant term, while background magnitude decreases through the crowding-out effect.

\textbf{Explanation-aware continual learning.} Recognizing that catastrophic forgetting manifests as explanation drift alongside accuracy
loss, \citet{ebrahimi2020remembering} introduced RRR, which stores Grad-CAM~\citep{selvaraju2017grad}
saliency maps alongside replay exemplars and penalizes attention shift, demonstrating that
models frequently preserve accuracy while fundamentally altering their decision process.
Subsequent work extends this paradigm: \citet{saha2023saliency} propose saliency-guided
exemplar packing to maximize buffer diversity, and \citet{liu2024task} introduce
task-adaptive saliency masks that adjust dynamically to forgetting patterns.
While these methods show that explanation consistency provides valuable supervision beyond classification loss, they lead to a \emph{self-referential drift loop}. This drift loop is not merely an attribution noise issue - it is a structural property of any scheme that derives targets from model-state-dependent saliency. The problem is compounded by well-documented brittleness of pixel-level attributions:
they are notoriously noisy and prone to highlighting spurious correlations~\citep{adebayo2018sanity,kindermans2019reliability}, can be sensitive to imperceptible input perturbations~\citep{ghorbani2019interpretation} and may emphasize high-frequency texture artifacts rather than semantically meaningful object parts~\citep{geirhos2018imagenet}.

\textbf{C$^3$L} breaks this loop, replacing model-state-dependent pixel targets with class-consistent, model-state-independent concept-level supervision generated entirely offline before training.

\textbf{Concept discovery and visual grounding.}
Early concept-based models required extensive manual annotation \cite{kim2018interpretability,koh2020concept}. Label-free CBMs \cite{oikarinen2023label} use LLMs to generate concepts without supervision. SAM~3 \cite{carion2025sam} enables precise spatial localization of textual concepts. In this work, we do not aim to construct a bottleneck model for inherent interpretability. Rather, we leverage these automated discovery and grounding pipelines as an offline pre-processing stage to synthesize rich, dense supervision signals. This allows us to anchor the model's attention to semantic regions during continual learning, effectively replacing expensive manual concept annotation with scalable, cross-instance semantic consistent priors.

\textbf{Positioning.} Prior region-level CL relies on model-state-dependent targets (e.g., CAM), creating a \textit{self-referential drift loop} we identify. \textbf{C$^3$L} eliminates this loop via three axes no prior method addresses simultaneously: (1)~\textit{target independence}: offline generation prevents drift reinforcement; (2)~\textit{semantic specificity}: supervision targets are fine-grained attributes rather than coarse boundaries; and (3)~\textit{cross-instance consistency}: class-shared targets induce robust invariants. Finally, Table~\ref{tab:concept_vs_object} isolates semantic specificity, confirming a significant advantage over object-level masks using the identical SAM~3 model.

\section{Method}
\label{sec:method}

\subsection{Problem Formulation}
We consider the class-incremental continual learning setting~\citep{rebuffi2017icarl}, where a model encounters a sequence of $T$ tasks $\{\mathcal{T}_1, \mathcal{T}_2, \ldots, \mathcal{T}_T\}$. Each task $\mathcal{T}_t$ provides a training set $\mathcal{D}_t = \{(x_i, y_i)\}_{i=1}^{N_t}$ containing $N_t$ labeled examples from $K_t$ new classes $\mathcal{C}_t = \{c_1^{(t)}, \ldots, c_{K_t}^{(t)}\}$ disjoint from previous tasks classes. During training on task $\mathcal{T}_t$, the model must learn to classify all classes encountered so far, $\mathcal{C}_{1:t} = \bigcup_{j=1}^t \mathcal{C}_j$, while only having access to $\mathcal{D}_t$ and a small replay buffer $\mathcal{M}$ containing stored samples from $\mathcal{D}_{1:t-1}$. Let $f_\theta: \mathcal{X} \rightarrow \mathcal{Y}$ denote our classifier and $E_\theta: \mathcal{X} \times \mathcal{Y} \rightarrow \mathbb{R}^{H \times W}$ an explanation function mapping input-label pairs to saliency maps. Beyond maximizing accuracy, we aim for \emph{semantic stability}: maintaining attention on semantically meaningful concept regions that define each class.

\textbf{Concept-grounded reasoning.} For each class $c \in \mathcal{C}_{1:t}$, let $\Psi_c = \{\psi_1, \psi_2, \ldots, \psi_K\}$ denote a set of visual concepts discovered for that class (\S\ref{sec:concept_discovery_grounding}). For sample $x$ with label $y$, let $M_{\text{concept}}^{(x)} \in \{0,1\}^{H \times W}$ denote regions where concepts from $\Psi_y$ are present (\S\ref{sec:concept_grounding}), and $M_{\text{bg}}^{(x)} = \mathbf{1} - M_{\text{concept}}^{(x)}$ its complement. Concept regions $M_{\text{concept}}^{(x)}$ identify meaningful features (e.g., wings, wheels) that should remain attended across tasks, while explanations in the $M_{\text{bg}}^{(x)}$ regions should not develop spurious correlations. By enforcing both positive supervision (attend to concepts) and negative supervision (stabilize backgrounds), we guide models toward focusing on robust, transferable features.

\subsection{Automated Concept Discovery and Grounding}
\label{sec:concept_discovery_grounding}

\textbf{Automatic concept generation.}
Manual concept annotation~\citep{koh2020concept} is impractical at the scale of continual learning. We adopt~\citet{oikarinen2023label}'s label-free approach, querying a large language model to generate discriminative textual attributes for each class $c \in \mathcal{C}_t$, yielding initial concepts $\Psi_c^{\text{raw}} = \{\psi_1, \ldots, \psi_K\}$ that are class-discriminative and linguistically interpretable. Large language models encode substantial knowledge about the visual appearance of object categories, enabling semantically meaningful attributes aligned with human perception.

\textbf{Concept filtering (Optional).}
We optionally validate concepts on held-out data $\mathcal{C}_c^{\text{VAL}}$ using: (1) \textbf{Occurrence rate} $\nu_{\text{occ}}(\psi) \geq \tau_{\text{occ}}$ - fraction of instances where concept $\psi$ can be spatially localized; (2) \textbf{Coverage} $\nu_{\text{cov}}$ - IoU between the union of all concept masks and the class-object segment, ensuring concepts are neither too sparse nor dense. Failing concepts are pruned, yielding $\Psi_c^* \subseteq \Psi_c^{\text{raw}}$. Importantly, Appendix~\ref{app:validation_ablation} shows this step is optional: unvalidated concept sets also achieve strong performance, indicating our framework tolerates some noise. Further details on the filtering mechanism can be found in the Appendix.

\textbf{Spatial concept grounding.}
\label{sec:concept_grounding}
For each replay sample $(x, y) \in \mathcal{M}$, we use SAM 3~\citep{carion2025sam} to localize concepts from $\Psi_y^*$. Given image $x$ and a batch of concept prompts (one noun phrase per concept), SAM 3 produces binary masks $\{B_i\}_{i=1}^M$ with confidence scores $\{s_i\}_{i=1}^M$ for each concept $\psi \in \Psi_y^*$. We aggregate high-confidence regions ($s_i \geq \tau_{\text{conf}}$):
\begin{equation}
M_{\text{concept}}^{(x)} = \bigcup_{\psi \in \Psi_y^*} \bigcup_{i: s_i(\psi) \geq \tau_{\text{conf}}} B_i^{(\psi)}.
\end{equation}
SAM 3's batched queries enable efficient single-pass grounding of all class concepts, eliminating the need for task-specific segmentation models or pixel-level annotations. The concept sets are shared across all instances of a class, and the masks are produced independently of the learner's state, ensuring cross-instance semantic consistency absent in pixel-level methods. Appendix~\ref{sec:abl_spatial_grounding} demonstrates that \method remains robust even under weak concept grounding conditions, as demonstrated by a targeted noise robustness evaluation.

\subsection{Concept-Consistent Continual Learning}
\label{sec:cccl}
\textbf{Explanation extraction.} We obtain pixel-level relevance $S_x \in \mathbb{R}^{H \times W}$ using architecture-specific methods satisfying relevance conservation~\citep{montavon2019layer}: Layer-wise Relevance Propagation (LRP)~\citep{bach2015pixel} with $\epsilon$-rule ($\epsilon = 10^{-6}$) for CNNs, and AttnLRP~\citep{achtibat2024attnlrp} for ViTs. LRP-based methods capture comprehensive layer-wise attribution. See Appendix~\ref{sec:abl_xai} for empirical comparison with different explanation methods.

\textbf{Training objective.} When training on task $\mathcal{T}_t$, we optimize:
\begin{equation}
\mathcal{L}_{\text{total}} = \mathcal{L}_{\text{task}} + \lambda_{\text{concept}} \mathcal{L}_{\text{concept}} + \lambda_{\text{stab}} \mathcal{L}_{\text{stab}}
\label{eq:total_loss}
\end{equation}
where $\mathcal{L}_{\text{task}}$ is cross-entropy on $\mathcal{D}_t$ and replay buffer $\mathcal{M}$.

The two regularization terms described below implement a \emph{correct-not-preserve} strategy, in contrast to conventional saliency alignment ($\lVert S - S^{\text{ref}} \rVert_1$) which mostly anchor the model to reference explanations that may themselves encode spurious correlations.

\textbf{Concept Overlap Preservation ($\mathcal{L}_{\text{concept}}$).}
To anchor attention to class-defining features, we enforce high relevance within concept regions:
\begin{equation}
\mathcal{L}_{\text{concept}} =  \frac{1}{|\mathcal{M}|} \sum_{(x,y) \in \mathcal{M}}\text{BCE}(S_x \odot M_{\text{concept}}^{(x)},  M_{\text{concept}}^{(x)})
\label{eq:concept_loss}
\end{equation}
where $\odot$ denotes element-wise multiplication, and BCE denotes the binary cross entropy function. Because concept targets are model-state-independent, they cannot encode drift, breaking the self-referential loop that afflicts pixel-level methods. Notably, $\mathcal{L}_{\text{concept}}$ can also be applied to all training samples, as discussed in Sec.~\ref{sec:main_results}.

\textbf{The Crowding-Out Effect.}
Minimizing $\mathcal{L}_{\text{concept}}$ also has an implicit effect on background regions. Let $S_x$ satisfy the approximate LRP conservation property~\citep{montavon2019layer}:
\begin{equation}
    \sum_{i \in \mathcal{C} \cup \mathcal{B}} S_x(i) \approx f_\theta(x)_y,
    \label{eq:conservation}
\end{equation}
where $\mathcal{C}$ and $\mathcal{B}$ index concept and background pixels respectively, and $f_\theta(x)_y$ is the logit of the target class. Denoting by $\rho_{\mathcal{C}} = \sum_{i \in \mathcal{C}} S_x(i) / \sum_{i} S_x(i)$ the share of relevance assigned to concept regions, Eq.~\ref{eq:conservation} gives $\sum_{i \in \mathcal{B}} S_x(i) \approx f_\theta(x)_y\,(1-\rho_{\mathcal{C}})$. Concept and background regions therefore compete for the relevance of the same prediction: increasing the concept share via $\mathcal{L}_{\text{concept}}$ reduces background relevance unless the target logit grows proportionally. We term this coupling the \emph{crowding-out effect}; it suppresses background relevance \emph{without explicit background penalization}. Since conservation is only approximate in networks with non-linearities, relevance is signed, and $\mathcal{L}_{\text{concept}}$ operates on min-max normalized maps, we treat crowding-out as a mechanism hypothesis rather than a guarantee, and validate it empirically in Appendix~\ref{sec:app_crowding}.

\textbf{Background Stability ($\mathcal{L}_{\text{stab}}$).}
The crowding-out effect reduces background relevance \emph{magnitude}, but does not control its \emph{spatial pattern}. Without additional regularization, freed relevance mass could drift toward new task-specific artifacts rather than
concentrating on diagnostic concepts. To prevent this, we store a reference saliency $S_x^{\text{ref}}$ when sample $(x,y)$ first enters $\mathcal{M}$ after task $\mathcal{T}_j$, and penalize spatial drift in background regions during subsequent tasks ($t > j$):
\begin{equation}
    \mathcal{L}_{\text{stab}} = \frac{1}{|\mathcal{M}|} \sum_{(x,y) \in \mathcal{M}}
    \left(1 - \frac{\langle S_x \odot M_{\text{bg}}^{(x)},\;
    S_x^{\text{ref}} \odot M_{\text{bg}}^{(x)} \rangle}
    {\|S_x \odot M_{\text{bg}}^{(x)}\|_2 \cdot
    \|S_x^{\text{ref}} \odot M_{\text{bg}}^{(x)}\|_2 + \epsilon}\right).
    \label{eq:stab_loss}
\end{equation}
Using cosine similarity is a deliberate design choice: it penalizes \emph{spatial and structural} drift in the background while leaving the magnitude free to decrease under crowding-out. Although $S_x^{\text{ref}}$ is model-derived, this term does not reintroduce the self-referential drift loop: it is confined to background regions and is scale-invariant, constraining \emph{where} background relevance lies but not \emph{how much} of it there is. The corrective signal comes from the model-independent concept term, which remains free to reduce background relevance, so spurious patterns present at buffer insertion cannot be reinforced in magnitude. Crucially, we stabilize rather than suppress backgrounds, since the pretrained backbone encodes robust contextual priors
(e.g., geometry, scale) that support generalization and should not be erased. Table~\ref{tab:abl_background} confirms this design outperforms removing background regularization.

A detailed discussion on objective design choices is provided in Appendix~\ref{sec:app_design_just}.

\textbf{Isolating the contribution of semantic guidance.} We rigorously verify that the observed gains stem from C$^3$L, rather than the capabilities of VLMs (Appendix~\ref{sec:app_foundation}). Specifically: (1) removing foundation models entirely by training on PartImageNet with ground-truth annotations - \method still outperforms all baselines; and (2) fine-grained concept masks outperform coarse object-level masks generated by the same SAM~3 model. Together, these results demonstrate that the semantic dual-regularization objective is essential to mitigating explanation drift.

\section{Experiments}
\label{sec:experiments}

We present a comprehensive empirical evaluation of \method across diverse continual learning scenarios, model architectures, and evaluation protocols. Our experiments address four key research questions:\\
 (1)~\textbf{RQ1} (\S\ref{sec:main_results}): Does concept-level regularization reduce catastrophic forgetting more effectively than state-of-the-art methods?\\
    (2)~\textbf{RQ2} (\S\ref{sec:explanation_quality}): Does \method mitigate spurious correlations and explanation drift?\\

    (3)~\textbf{RQ3} (\S\ref{sec:ablations}): What are the contributions of C$^3$L's key components and design choices - dual regularization, buffer size, spatial grounding, explanation methods, task granularity, and concept discovery mechanism?\\
    (4)~\textbf{RQ4} (\S\ref{subsec:robust}): Can C$^3$L's training mechanism improve model robustness to out-of-distribution shifts (ImageNet-A and ObjectNet)? \textbf{Success here suggests
the model has internalized and generalized transferable semantic concepts rather than dataset-specific artifacts.}

\subsection{Experimental Setup}
\label{sec:setup}

\textbf{Datasets.} We evaluate on five benchmarks spanning fine-grained recognition and out-of-distribution generalization: CUB-200-2011 (\textbf{CUB})~\citep{wah2011caltech} (200 bird species), \textbf{DomainNet}~\citep{peng2019moment} (345 classes across 6 domains), \textbf{ImageNet-R}~\citep{hendrycks2021many} (200 classes with artistic renditions), \textbf{ImageNet-A}~\citep{hendrycks2021natural} (200 classes with natural adversarial examples), \textbf{ObjectNet}~\citep{barbu2019objectnet} (113 classes with pose/background variations). These datasets represent both typical class-incremental benchmarks (CUB, DomainNet) and challenging distribution shifts (ImageNet-R/A, ObjectNet) that test robustness beyond standard continual learning settings. Also, they exhibit a significant domain gap with ImageNet (i.e., the pre-trained dataset). Following standard protocols~\citep{buzzega2020dark,wang2022dualprompt}, we split datasets into 10 sequential tasks (5 for DomainNet) with equal classes per task under class-incremental learning: task boundaries known during training but not at test, with classification over all encountered classes. For ObjectNet, 3 randomly sampled classes are excluded so that the remaining 110 classes divide evenly into tasks (Appendix~\ref{sec:app_implementation}).

\textbf{Evaluation metrics.} We report three standard continual learning metrics~\citep{rebuffi2017icarl,chaudhry2018riemannian}: (i) \textbf{Average Accuracy (ACC)} $\uparrow$: the average accuracy on all the tasks after the last task has been learned, (ii) \textbf{Average Forgetting (F)} $\downarrow$: mean accuracy drop on previous tasks, $F = \frac{1}{T-1}\sum_{i=1}^{T-1} \max_{j \in \{i,\ldots,T\}} A_{i,j} - A_{i,T}$, where $A_{i,j}$ is accuracy on task $i$ after training on task $j$, (iii) \textbf{Average Incremental Accuracy (AIA)} $\uparrow$: mean accuracy after learning each task, capturing the model's performance throughout the entire learning trajectory. For explanation evaluation (\S\ref{sec:explanation_quality}), we measure: (i)~\textbf{Spurious Correlation Mitigation (Debiasing)}:
following~\citet{lee2024continual} and~\citet{bayasi2024biaspruner} we evaluate ACC over both Fitzpatrick17K (\textbf{FITZ})~\cite{bayasi2024biaspruner} and Split ImageNet-100C (\textbf{I-100C})~\cite{lee2024continual} to evaluate the model's reliance on spurious correlations.
(ii)~\textbf{Semantic Localization}: following~\citet{chefer2021transformer} we measure the mIoU, pixel accuracy (pixAcc), and mAP of the explanations on CUB. More details provided in Appendix~\ref{sec:app_xai_qual}.

\textbf{Baselines.} We compare against nearly 20 state-of-the-art methods spanning multiple paradigms. For CNNs: \emph{Regularization-based}: EWC~\citep{kirkpatrick2017overcoming}, LwF~\citep{li2016learning}, PEC~\citep{kim2024prediction}; \emph{Replay-based}: ER~\citep{riemer2018learning}, iCaRL~\citep{rebuffi2017icarl}, DER++~\citep{buzzega2020dark}, ER-ACE~\citep{caccia2022new}, CSReL-CL~\citep{park2024coreset};
\emph{Explanation-aware}: RRR~\citep{ebrahimi2020remembering}, EPR~\citep{saha2023saliency} TASS~\citep{liu2024task}. For ViTs: \emph{Prompt-based}: L2P~\citep{wang2022learning}, DualPrompt~\citep{wang2022dualprompt}, CODA-Prompt~\citep{smith2023coda}; \emph{Adapter-based}: MOS~\citep{chen2025mos}, MIN~\citep{wang2025min}, PLAN~\citep{zhao2025plan}. We also provide a ViT-based implementation for RRR named RRR-ViT which uses Grad-CAM-ViT~\cite{chefer2021transformer}. Joint Training is standard supervised fine-tuning on the i.i.d. union of all tasks' data, without concept supervision; it is commonly regarded as an upper bound for continual learning methods that use the same supervision. In this work, we focus on continual learning for widely used architectures such as CNNs and ViTs. While \method could potentially be extended to concept bottleneck models (CBMs), systematically adapting CBMs to continual learning constitutes a research problem in its own and is therefore left for future work.

\textbf{Controlled comparisons.} To isolate the benefit of the choice of explanation method from C$^3$L, we implement two controlled variants of RRR, called RRR-LRP and RRR-ViT-AttnLRP (for ResNet-18 and ViT-B respectively), that use our explanation methods instead of Grad-CAM. To isolate replay benefits, we add replay-augmented variants (-R) following~\citet{wang2022learning}.

All replay methods use identical buffer size ($|\mathcal{M}| = 1000$) for fair comparison. Section~\ref{sec:ablations} shows that \method achieves competitive results against all baselines (with $|\mathcal{M}| = 1000$) using a small buffer of size $|\mathcal{M}| = 200$.
We follow original code-bases for implementation and hyperparameter settings to ensure the best possible performance.

\textbf{Implementation Details.} We use ResNet-18~\citep{he2016deep} (ImageNet-1K pretrained) and ViT-B/16~\citep{dosovitskiy2020image} (ImageNet-21K pretrained). \method employs LRP~\citep{bach2015pixel} for CNNs and AttnLRP~\citep{achtibat2024attnlrp} for ViTs; all baselines use Grad-CAM~\citep{selvaraju2017grad} as a unified evaluation protocol. Concept discovery follows~\citet{oikarinen2023label} with spatial grounding via SAM 3~\citep{carion2025sam}. Full configuration details are provided in Appendix~\ref{sec:app_implementation}.

\textbf{Training Details.} Regularization is applied exclusively to replay samples ($|\mathcal{M}| = 1000$). Results are averaged over 5 runs (mean $\pm$ std), with statistical significance assessed via paired t-tests ($p < 0.05$). Full hyperparameter settings and training protocols are provided in Appendix~\ref{sec:app_implementation}.

Finally, comprehensive implementation details are provided in the Appendix. Across all tables, the best results are in \textbf{bold}, and the second-best are \underline{underlined}.

\subsection{RQ1: Standard Continual Learning Performance}
\label{sec:main_results}

\begin{table*}[t]
\centering

\caption{Quantitative comparison of Continual Learning performance across five benchmarks using ResNet-18 and ViT-B architectures. We report Average Accuracy (ACC), Average Forgetting (F), and Average Incremental Accuracy (AIA). We report results over 5 trials.}

\label{tab:results_cnn}
\resizebox{\textwidth}{!}{
\begin{tabular}{l|ccc|ccc|ccc|ccc|ccc}
\toprule
& \multicolumn{3}{c|}{{CUB}} & \multicolumn{3}{c|}{{DomainNet}} & \multicolumn{3}{c|}{{ImageNet-R}} & \multicolumn{3}{c|}{{ImageNet-A}} & \multicolumn{3}{c}{{ObjectNet}} \\
\textbf{Method} & ACC $\uparrow$ & F $\downarrow$ & AIA $\uparrow$ & ACC $\uparrow$ & F $\downarrow$ & AIA $\uparrow$ & ACC $\uparrow$ & F $\downarrow$ & AIA $\uparrow$ & ACC $\uparrow$ & F $\downarrow$ & AIA $\uparrow$ & ACC $\uparrow$ & F $\downarrow$ & AIA $\uparrow$ \\
\midrule
\multicolumn{16}{c}{\emph{ResNet-18}}\\
\midrule
Joint Training & 83.04${\pm 0.51}$ & - & - & 74.99${\pm 0.22}$ & - & - & 76.31${\pm 0.90}$ & - & - & 71.22${\pm 0.42}$ & - & - & 73.42${\pm 0.91}$ & - & - \\
\midrule
\multicolumn{16}{l}{\emph{Replay-based}} \\
ER & 64.34${\pm 0.92}$ & 14.25${\pm 0.58}$ & 68.42${\pm 0.84}$ & 61.92${\pm 0.81}$ & 8.92${\pm 0.55}$ & 61.99${\pm 0.76}$ & 63.38${\pm 0.68}$ & 11.85${\pm 0.49}$ & 66.15${\pm 0.72}$ & 55.01${\pm 0.45}$ & 19.43${\pm 0.66}$ & 59.80${\pm 0.38}$ & 57.84${\pm 0.52}$ & 17.24${\pm 0.58}$ & 62.01${\pm 0.55}$ \\
DER++ & 71.45${\pm 0.72}$ & 9.62${\pm 0.45}$ & 75.14${\pm 0.65}$ & 68.02${\pm 0.58}$ & 5.74${\pm 0.32}$ & 67.70${\pm 0.55}$ & 69.81${\pm 0.48}$ & 7.82${\pm 0.35}$ & 72.15${\pm 0.52}$ & 61.43${\pm 0.38}$ & 12.53${\pm 0.44}$ & 66.41${\pm 0.32}$ & 64.44${\pm 0.42}$ & 11.34${\pm 0.48}$ & 68.80${\pm 0.45}$ \\
ER-ACE & 69.73${\pm 0.78}$ & 10.85${\pm 0.49}$ & 73.32${\pm 0.72}$ & 66.44${\pm 0.65}$ & 6.42${\pm 0.38}$ & 66.22${\pm 0.62}$ & 68.21${\pm 0.55}$ & 8.93${\pm 0.42}$ & 70.75${\pm 0.58}$ & 59.83${\pm 0.42}$ & 14.82${\pm 0.52}$ & 64.71${\pm 0.36}$ & 62.74${\pm 0.46}$ & 13.14${\pm 0.52}$ & 67.11${\pm 0.48}$ \\
CSReL-CL & 71.92${\pm 0.82}$ & 9.25${\pm 0.52}$ & 75.54${\pm 0.76}$ & 67.44${\pm 0.69}$ & 6.12${\pm 0.42}$ & 67.23${\pm 0.65}$ & 69.42${\pm 0.59}$ & 8.24${\pm 0.46}$ & 71.83${\pm 0.62}$ & 60.74${\pm 0.45}$ & 13.42${\pm 0.56}$ & 65.51${\pm 0.39}$ & 63.83${\pm 0.49}$ & 12.05${\pm 0.56}$ & 67.94${\pm 0.53}$ \\
\midrule
\multicolumn{16}{l}{\emph{Explanation-aware}} \\
RRR & 72.21${\pm 0.78}$ & 9.10${\pm 0.44}$ & 75.81${\pm 0.73}$ & 68.25${\pm 0.33}$ & 5.66${\pm 1.06}$ & 67.95${\pm 0.54}$ & 70.13${\pm 0.65}$ & 7.66${\pm 0.62}$ & 72.43${\pm 0.50}$ & 61.73${\pm 0.43}$ & 12.06${\pm 0.21}$ & 66.68${\pm 0.31}$ & 64.82${\pm 0.32}$ & 10.87${\pm 0.35}$ & 69.16${\pm 0.64}$ \\
RRR-LRP & \underline{73.25${\pm 0.65}$} & \underline{8.42${\pm 0.38}$} & \underline{76.74${\pm 0.58}$} & \underline{69.30${\pm 0.52}$} & \underline{5.13${\pm 1.28}$} & \underline{68.92${\pm 0.49}$} & \underline{71.15${\pm 0.45}$} & \underline{6.92${\pm 0.92}$} & \underline{73.32${\pm 0.46}$} & \underline{63.01${\pm 0.35}$} & \underline{10.24${\pm 0.69}$} & \underline{68.12${\pm 0.28}$} & \underline{66.20${\pm 0.36}$} & \underline{9.23${\pm 0.42}$} & \underline{70.74${\pm 0.41}$} \\
EPR & 72.86${\pm 0.72}$ & 8.75${\pm 0.41}$ & 76.32${\pm 0.65}$ & 68.82${\pm 0.42}$ & 5.36${\pm 0.82}$ & 68.41${\pm 0.51}$ & 70.68${\pm 0.55}$ & 7.21${\pm 0.84}$ & 72.88${\pm 0.48}$ & 62.45${\pm 0.39}$ & 11.15${\pm 0.52}$ & 67.42${\pm 0.30}$ & 65.54${\pm 1.04}$ & 10.02${\pm 0.38}$ & 70.05${\pm 0.52}$ \\
TASS & 70.24${\pm 0.85}$ & 10.35${\pm 0.56}$ & 74.12${\pm 0.78}$ & 66.91${\pm 0.72}$ & 6.74${\pm 0.45}$ & 66.73${\pm 0.68}$ & 68.92${\pm 0.62}$ & 8.63${\pm 0.49}$ & 71.42${\pm 0.65}$ & 60.13${\pm 0.48}$ & 14.25${\pm 0.62}$ & 65.01${\pm 0.42}$ & 63.01${\pm 0.52}$ & 12.83${\pm 0.58}$ & 67.44${\pm 0.55}$ \\
\midrule
\method  & $\mathbf{79.73{\pm 0.31}}$ & $\mathbf{5.32{\pm 0.63}}$ & $\mathbf{82.24{\pm 0.35}}$ & $\mathbf{73.01{\pm 0.23}}$ & $\mathbf{3.14{\pm 0.55}}$ & $\mathbf{72.51{\pm 0.88}}$ & $\mathbf{74.35{\pm 0.46}}$ & $\mathbf{4.23{\pm 0.29}}$ & $\mathbf{75.84{\pm 0.65}}$ & $\mathbf{67.84{\pm 0.72}}$ & $\mathbf{6.52{\pm 0.46}}$ & $\mathbf{70.83{\pm 0.41}}$ & $\mathbf{71.12{\pm 0.53}}$ & $\mathbf{5.82{\pm 0.25}}$ & $\mathbf{72.23{\pm 0.92}}$ \\
\midrule
\multicolumn{16}{c}{\emph{ViT-B}}\\
\midrule
Joint Training & 86.12${\pm 0.23}$ & - & - & 81.31${\pm 0.29}$ & - & - & 84.59${\pm 0.31}$ & - & - & 75.31${\pm 0.19}$ & - & - & 78.96${\pm 0.24}$ & - & - \\
\midrule
\multicolumn{16}{l}{\emph{Prompt-based}} \\
L2P & 70.82${\pm 1.24}$ & 7.94${\pm 0.65}$ & 74.24${\pm 0.91}$ & 71.85${\pm 0.78}$ & 4.23${\pm 0.42}$ & 72.40${\pm 0.83}$ & 71.49${\pm 0.69}$ & 4.85${\pm 0.56}$ & 71.91${\pm 0.74}$ & 54.94${\pm 0.82}$ & 6.83${\pm 0.44}$ & 56.11${\pm 0.57}$ & 64.53${\pm 0.70}$ & 6.24${\pm 0.55}$ & 66.12${\pm 0.69}$ \\
CODA-Prompt & 74.74${\pm 0.79}$ & 5.42${\pm 0.42}$ & 78.13${\pm 0.67}$ & 74.95${\pm 0.82}$ & 3.12${\pm 0.38}$ & 75.50${\pm 0.78}$ & 72.35${\pm 0.65}$ & 3.82${\pm 0.41}$ & 72.69${\pm 0.64}$ & 54.24${\pm 0.68}$ & 6.92${\pm 0.40}$ & 55.49${\pm 0.53}$ & 67.82${\pm 0.66}$ & 4.83${\pm 0.42}$ & 68.24${\pm 0.55}$ \\
\midrule
\multicolumn{16}{l}{\emph{Adapter-based}} \\
MOS & 75.94${\pm 0.68}$ & 4.82${\pm 0.41}$ & 79.53${\pm 0.55}$ & 76.14${\pm 0.70}$ & 2.12${\pm 0.39}$ & 76.81${\pm 0.66}$ & 76.88${\pm 0.24}$ & 4.15${\pm 0.61}$ & 78.35${\pm 0.53}$ & 61.34${\pm 0.68}$ & 3.52${\pm 0.29}$ & 62.26${\pm 0.35}$ & 71.53${\pm 0.67}$ & 3.54${\pm 0.31}$ & 71.82${\pm 0.56}$ \\
PLAN & 76.12${\pm 0.70}$ & 4.83${\pm 0.42}$ & 79.74${\pm 0.56}$ & 76.03${\pm 0.83}$ & 2.24${\pm 0.28}$ & 76.76${\pm 0.67}$ & 76.62${\pm 0.55}$ & 4.02${\pm 0.31}$ & 78.93${\pm 0.84}$ & 60.25${\pm 0.69}$ & 4.52${\pm 0.32}$ & 61.14${\pm 0.55}$ & 70.82${\pm 0.68}$ & 3.42${\pm 0.30}$ & 71.23${\pm 1.03}$ \\
\midrule
\multicolumn{16}{l}{\emph{Replay-based}} \\
RRR-ViT & 76.94${\pm 1.17}$ & 4.82${\pm 0.61}$ & 78.53${\pm 1.14}$ & 78.32${\pm 1.21}$ & 2.01${\pm 0.49}$ & 78.19${\pm 1.16}$ & 78.83${\pm 0.94}$ & 4.12${\pm 0.50}$ & 80.84${\pm 0.93}$ & 60.52${\pm 1.18}$ & 5.43${\pm 0.23}$ & 63.24${\pm 0.64}$ & 70.19${\pm 1.17}$ & 3.82${\pm 0.51}$ & 70.94${\pm 1.16}$ \\
RRR-ViT-AttnLRP & 78.09${\pm 0.85}$ & 4.17${\pm 0.58}$ & 80.34${\pm 0.88}$ & 78.47${\pm 1.08}$ & 1.92${\pm 0.37}$ & 78.56${\pm 0.90}$ & 79.48${\pm 0.76}$ & 3.35${\pm 1.01}$ & 80.87${\pm 0.66}$ & 63.02${\pm 0.56}$ & 4.24${\pm 0.26}$ & 63.51${\pm 0.59}$ & 71.61${\pm 0.87}$ & 3.33${\pm 0.42}$ & 71.88${\pm 0.89}$ \\
L2P-R & 69.62${\pm 1.25}$ & 6.34${\pm 0.56}$ & 77.25${\pm 0.92}$ & 74.23${\pm 0.29}$ & 3.42${\pm 0.53}$ & 74.82${\pm 1.04}$ & 76.82${\pm 0.90}$ & 3.91${\pm 0.57}$ & 77.34${\pm 0.89}$ & 58.33${\pm 0.23}$ & 6.14${\pm 0.21}$ & 59.71${\pm 0.68}$ & 67.24${\pm 1.21}$ & 5.32${\pm 0.56}$ & 68.15${\pm 0.61}$ \\
CODA-R & 77.53${\pm 0.99}$ & 4.23${\pm 0.42}$ & 80.24${\pm 0.77}$ & 78.02${\pm 0.82}$ & 3.02${\pm 0.28}$ & 78.21${\pm 0.78}$ & 78.54${\pm 0.66}$ & 2.82${\pm 0.31}$ & 79.52${\pm 0.64}$ & 61.24${\pm 0.89}$ & 4.82${\pm 0.42}$ & 61.43${\pm 0.55}$ & 71.53${\pm 0.78}$ & 3.24${\pm 0.30}$ & 71.62${\pm 0.67}$ \\
MOS-R & 78.82${\pm 0.65}$ & 3.84${\pm 0.29}$ & \underline{82.52${\pm 0.53}$} & \underline{79.03${\pm 0.78}$} & \underline{1.52${\pm 0.42}$} & \underline{79.42${\pm 0.54}$} & \underline{80.92${\pm 0.42}$} & \underline{2.43${\pm 0.28}$} & \underline{81.55${\pm 0.31}$} & \underline{67.84${\pm 0.96}$} & \underline{2.24${\pm 0.27}$} & \underline{66.25${\pm 0.42}$} & \underline{73.62${\pm 0.35}$} & \underline{2.52${\pm 0.29}$} & \underline{73.89${\pm 0.93}$} \\
PLAN-R & \underline{79.24${\pm 0.56}$} & \underline{3.52${\pm 0.28}$} & 82.14${\pm 0.54}$ & 78.62${\pm 0.59}$ & 1.83${\pm 0.27}$ & 78.92${\pm 0.65}$ & {80.13${\pm 0.51}$} & 2.58${\pm 1.29}$ & {80.89${\pm 0.42}$} & 65.48${\pm 0.57}$ & 3.05${\pm 0.31}$ & 65.83${\pm 0.51}$ & 72.85${\pm 0.56}$ & 2.84${\pm 0.28}$ & 73.06${\pm 0.55}$ \\
\midrule
\method  & $\mathbf{83.65{\pm 0.28}}$ & $\mathbf{2.92{\pm 0.24}}$ & $\mathbf{84.64{\pm 0.37}}$ & $\mathbf{80.43{\pm 0.55}}$ & $\mathbf{1.23{\pm 0.62}}$ & $\mathbf{80.82{\pm 0.26}}$ & $\mathbf{82.74{\pm 0.76}}$ & $\mathbf{1.89{\pm 0.33}}$ & $\mathbf{83.92{\pm 0.45}}$ & $\mathbf{71.14{\pm 0.59}}$ & $\mathbf{1.53{\pm 0.41}}$ & $\mathbf{70.93{\pm 0.38}}$ & $\mathbf{76.83{\pm 0.56}}$ & $\mathbf{1.92{\pm 0.83}}$ & $\mathbf{77.92{\pm 0.78}}$ \\
\bottomrule
\end{tabular}
}
\end{table*}

Table~\ref{tab:results_cnn} presents comprehensive results across ResNet-18 and ViT-B. Due to space constraints, the complete set of results, including all evaluated methods, is reported on Table~\ref{tab:results_cnn_full} in the Appendix. \method achieves consistent improvements in accuracy and forgetting
across all datasets, outperforming state-of-the-art and explanation-based methods. Notably, $\mathcal{L}_{\text{concept}}$ could also be applied to all training samples. However, we restrict it to replay examples to ensure a fair comparison with prior replay methods. Results for this variant are reported in the full results table in the Appendix.

A critical question is whether gains stem from using superior explanation methods (LRP vs. Grad-CAM) rather than concept-level regularization. Our controlled comparisons provide clear evidence for the latter. RRR-ViT-AttnLRP (using AttnLRP instead of Grad-CAM) improves over RRR-ViT. However, \method with AttnLRP achieves far greater gains. Since both use identical AttnLRP, this provides evidence for the benefit of concept-level over pixel-level regularization: semantically grounded regions provide substantially stronger supervision than raw saliency. The same pattern holds for ResNet-18.

Replay-based methods mostly outperform other methods in both CNNs and ViTs. This aligns with recent findings~\cite{wan2025understanding} showing that replay-based methods benefit from preserving feature learning dynamics. We further observe that for ResNet-18, explanation-aware methods achieve very good performance, and are mostly better than regular replay-based methods. However, pixel-level methods may not distinguish between spurious correlations and semantic features. In contrast, \method focuses on validated semantic concepts, providing cleaner supervision, which is also evident by the performance. This gap is further emphasized on Sec.~\ref{sec:explanation_quality}. ViTs exhibit substantially lower forgetting than CNNs across all methods, likely reflecting the more stable representations of large-scale pretrained transformers.
Notably, RRR-ViT sometimes outperforms prompt-based methods with replay (L2P-R, CODA-R), demonstrating that explanation consistency provides valuable regularization even for transformers. Nevertheless, \method maintains consistent advantages across both architectures, demonstrating its effectiveness across distinct architectural families.

\subsection{RQ2: Explanation Drift and Spurious Correlations}
\label{sec:explanation_quality}
Standard continual learning protocols do not assess whether methods mitigate spurious correlations or maintain focus on semantically meaningful features. To address this limitation, we adopt two complementary evaluation protocols:
(1)~\textbf{Spurious Correlation Mitigation}, measured by accuracy on FITZ and I100C, evaluating model reliance on spurious features (skin tone in FITZ and injected noise in I-100C); (2)~\textbf{Semantic Localization}, evaluated using mIoU, pixAcc, and mAP of explanations on CUB annotated segments\footnote{\url{https://data.caltech.edu/records/w9d68-gec53}}, to quantify explanation drift. For (1), we follow prior work~\cite{bayasi2024biaspruner, lee2024continual} and fine-tune a CNN model (ResNet-18 pretrained on ImageNet-1K). For (2), we evaluate the same CUB-finetuned model from Section~\ref{sec:main_results}. We compare \method against representative baselines from Table~\ref{tab:results_cnn} and Table~\ref{tab:results_cnn_full}. In addition, for protocol (1), we include two recent class-incremental learning methods for debiasing: BiasPruner (\textbf{BP})~\cite{bayasi2024biaspruner} and \textbf{BGS}~\cite{lee2024continual}. Further experimental and implementation details provided in Appendix~\ref{sec:app_xai_qual}. As shown in Table~\ref{tab:segmentation}, \method consistently achieves the best performance across both accuracy and semantic localization metrics. The improvement in accuracy is consistent with our hypothesis that \method anchors attention to semantically relevant features, thereby preventing the emergence of spurious correlations. Notably, \method outperforms specialized debiasing methods on accuracy across both datasets. In contrast, pixel-level explanation-aware baselines (RRR, EPR) and other continual learning methods (L2P, MOS, MIN) show lower semantic localization scores and spurious correlation mitigation performance. These results show that concept-grounded regularization systematically redirects model attention from irrelevant cues to semantically meaningful features, thereby mitigating explanation drift, improving the quality of model reasoning, and limiting the accumulation of spurious correlations.

\begin{table}[h]
    \centering
    \small
    \begin{minipage}{0.45\textwidth}
        \centering
        \caption{Evaluation of explanation quality and stability using segmentation metrics (mIoU, pixAcc, mAP) on the CUB dataset and ACC on the FITZ and the I-100C datasets. Higher is better for all metrics.}
        \vspace{2mm}
        \label{tab:segmentation}
        \resizebox{\linewidth}{!}{%
        \begin{tabular}{l|ccccc}
        \toprule
        \multicolumn{6}{c}{\emph{ResNet-18}} \\
        \textbf{Metric} & {RRR} & {EPR} & {BGS} & {BP} & \textbf{\method{}} \\
        \midrule
        {ACC~(FITZ)}    & 38.69 & 40.58 & 56.85 & 58.13 & \textbf{60.47} \\
        {ACC~(I-100C)}    & 31.49 & 32.97 & 51.39 & 48.52 & \textbf{64.19} \\
        {mIoU}   & 52.59 & 53.83 & 49.65 & 50.24 & \textbf{58.77} \\
        {pixAcc} & 66.92 & 67.40 & 63.22 & 62.47 & \textbf{79.96} \\
        {mAP}    & 69.71 & 68.32 & 65.89 & 61.51 & \textbf{83.92} \\
        \midrule
        \multicolumn{6}{c}{\emph{ViT-B}} \\
        \textbf{Metric} & {RRR-ViT} & {L2P} & {MOS} & {MIN} & \textbf{\method{}} \\
        \midrule
        {mIoU}   & 54.55 & 42.02 & 54.17 & 51.39 & \textbf{59.86} \\
        {pixAcc} & 72.05 & 56.15 & 71.45 & 62.23 & \textbf{80.51} \\
        {mAP}    & 68.28 & 43.59 & 67.86 & 55.43 & \textbf{85.83} \\
        \bottomrule
        \end{tabular}
        }
    \end{minipage}
    \hfill
    \begin{minipage}{0.53\textwidth}
        \centering
        \caption{Ablation study analyzing the impact of removing concept regularization ($\mathcal{L}_{concept}$) and background stability ($\mathcal{L}_{stab}$) terms.}
        \vspace{2mm}
        \label{tab:ablations}
        \resizebox{\linewidth}{!}{
        \begin{tabular}{l|ccc|ccc}
        \toprule
        \multirow{2}{*}{\textbf{Configuration}} & \multicolumn{3}{c|}{\textbf{ImageNet-R}} & \multicolumn{3}{c}{\textbf{CUB}} \\
         & ACC $\uparrow$ & F $\downarrow$ & AIA $\uparrow$ & ACC $\uparrow$ & F $\downarrow$ & AIA $\uparrow$ \\
        \midrule
        \method{} & 82.74 & 1.89 & 83.92 & 83.65 & 2.92 & 84.64 \\
        \midrule
        w/o $\mathcal{L}_{\text{concept}}$ & 77.12 & 3.82 & 78.13 & 73.74 & 5.19 & 72.95 \\
        w/o $\mathcal{L}_{\text{stab}}$ & 78.94 & 3.01 & 78.32 & 78.42 & 3.86 & 78.98 \\
        w/o both & 72.27 & 9.13 & 71.56 & 70.92 & 8.48 & 71.40 \\
        \bottomrule
        \end{tabular}
        }
    \end{minipage}
\end{table}

\vspace{-3mm}
\subsection{RQ3: Ablation Studies}
\label{sec:ablations}

We dissect C$^3$L's components to understand their individual contributions. All ablations use ImageNet-R with ViT-B unless otherwise noted. We address: (1) the benefit of each regularization term, (2) impact of replay buffer size, (3) concept-level vs. object-level segmentation (\S\ref{sec:abl_concept_source}) - \textbf{ emphasizes the benefit from using concept guidance}, (4) alternative explanation methods (\S\ref{sec:abl_xai}), (5) task granularity (\S\ref{sec:task_splitting}), (6) effect of replay vs. no-replay training using concepts (\S\ref{sec:abl_replay_analysis}), and (7) SAM 3 vs. Grounded SAM (\S\ref{sec:abl_spatial_grounding}). Space constraints limit the main paper to questions (1) and (2). The remaining ablations appear in the Appendix~\ref{sec:app_additional_abl}.

\textbf{Regularization terms.}
Table~\ref{tab:ablations} shows results for component ablations. On ImageNet-R, both terms cause similar performance drops, indicating both are important. On CUB, removing $\mathcal{L}_{\text{concept}}$ causes larger drops, suggesting this term is especially impactful for fine-grained datasets by strengthening concept attention. Yet, removing $\mathcal{L}_{\text{stab}}$ also hurts performance, confirming its role preventing drift to spurious backgrounds and textures that correlate with specific tasks. The asymmetric importance confirms our design intuition: concept overlap provides positive supervision, while stability provides negative supervision.

\textbf{Impact of Replay Buffer Size.}
We evaluate \method on ImageNet-R using the ViT-B backbone with varying replay buffer sizes $|\mathcal{M}| \in \{200, 500, 1000\}$. Standard replay methods degrade severely with smaller buffers due to overfitting and reduced representative variance. As presented in Table~\ref{tab:buffer_size}, \method demonstrates robustness to reduced replay capacity. With the standard buffer size of $|\mathcal{M}|=1000$, \method consistently outperforms all baselines. Notably, even when the buffer size is reduced to $|\mathcal{M}|=500$, our method maintains superior performance, surpassing all compared methods, even those using the full 1,000-sample buffer. In the highly constrained regime of $|\mathcal{M}|=200$, \method remains highly competitive. While the performance gap naturally narrows, our method frequently matches and even exceeds the performance of strong baselines trained with larger buffers. This data efficiency is consistent with our hypothesis: concept-grounded regularization anchors the model to semantic features which can be more robust and transferable than the pixel-level patterns relied upon by standard replay and saliency methods, thus requiring fewer examples to maintain stability.

\begin{table}[h]
    \small
    \centering
    \begin{minipage}{0.59\textwidth}
        \centering
        \caption{Performance analysis on ImageNet-R (ViT-B) with varying replay buffer sizes ($|\mathcal{M}|$) to assess data efficiency.}
        \vspace{2mm}
        \label{tab:buffer_size}
        \resizebox{\linewidth}{!}{%
        \begin{tabular}{l|ccc|ccc|ccc}
            \toprule
            \multirow{2}{*}{\textbf{Method}} & \multicolumn{3}{c|}{$|\mathcal{M}|=$\textbf{200}} & \multicolumn{3}{c|}{$|\mathcal{M}|=$\textbf{500}} & \multicolumn{3}{c}{$|\mathcal{M}|=$\textbf{1,000}} \\
             & ACC $\uparrow$ & F $\downarrow$ & AIA $\uparrow$ & ACC $\uparrow$ & F $\downarrow$ & AIA $\uparrow$ & ACC $\uparrow$ & F $\downarrow$ & AIA $\uparrow$ \\
            \midrule
            RRR-ViT & 71.48 & 7.25 & 70.20 & 73.83 & 6.11 & 73.55 & 78.83 & 4.12 & 80.84 \\
            L2P-R  & 72.83 & 4.44 & 72.98 & 73.60 & 4.18 & 76.32 & 76.82 & 3.91 & 77.34 \\
            CODA-R    & 73.80 & 3.43 & 73.41 & 75.13 & 3.16 & 77.25 & 78.54 & 2.82 & 79.52 \\
            MOS-R            & 77.28 & 3.88 & 78.95 & 78.57 & 3.21 & 78.92 & 80.92 & 2.43 & 81.55 \\
            PLAN-R         & 77.78 & 3.45 & 79.65 & 78.23 & 3.18 & 80.11 & 80.13 & 2.58 & 80.89 \\
            \midrule
            \textbf{\method}                  & \textbf{80.40} & \textbf{2.65} & \textbf{81.89} & \textbf{81.26} & \textbf{2.11} & \textbf{82.99} & \textbf{82.74} & \textbf{1.89} & \textbf{83.92} \\
            \bottomrule
        \end{tabular}
        }
    \end{minipage}
    \hfill
    \begin{minipage}{0.39\textwidth}
        \centering
        \caption{Out-of-distribution robustness evaluation. We report accuracy before and after fine-tuning. For ResNet-18 we compared: RRR~/~\method, and for ViT-B: RRR-ViT~/~\method.}
        \vspace{2mm}
        \label{tab:robustness_results}
        \resizebox{\linewidth}{!}{%
        \begin{tabular}{llcc}
            \toprule
            {Model} & {Dataset} & {Before} & {After} \\
            \midrule
            \multirow{2}{*}{{ResNet-18}}
            & ImageNet-A & 8.59 & 7.92~/~\textbf{ 20.15} \\
            & ObjectNet  & 21.88 & 21.56~/~\textbf{ 36.93} \\
            \midrule
            \multirow{2}{*}{{ViT-B}}
            & ImageNet-A & 13.48 & 13.18~/~\textbf{30.82} \\
            & ObjectNet  & 33.92 & 34.49~/~\textbf{58.04} \\
            \bottomrule
        \end{tabular}
        }
    \end{minipage}
\end{table}

\vspace{-3mm}
\subsection{RQ4: Robustness to Out-of-Distribution (OOD) Samples}
\label{subsec:robust}
We evaluate robustness to OOD samples using a setup inspired by~\cite{yosinski2014transferable}, which studies feature transferability by training on a subset of ImageNet classes. Specifically, we construct a compact fine-tuning dataset for \method by randomly selecting 5 images per class from half of the ImageNet-1K~\cite{deng2009imagenet} categories, resulting in a total of 2,500 training images. This sparse sampling is motivated by computational considerations and serves to assess data efficiency under limited supervision, while random class selection ensures broad semantic coverage. Experiments are conducted using ResNet-18 and ViT-B models pretrained on ImageNet-1K. Robustness is evaluated specifically on ImageNet-A and ObjectNet (restricted to classes overlapping  ImageNet-1K) as these benchmarks contain challenging samples that commonly cause failures in ImageNet-pretrained models. Further details are provided in the Appendix. Table~\ref{tab:robustness_results} details the classification accuracy before and after fine-tuning with \method in comparison with RRR (RRR-ViT). We observe a significant increase in accuracy when employing C$^3$L's semantic-based regularization. In contrast, RRR yields only negligible gains (and sometimes even decrease the accuracy), highlighting its lower robustness capabilities in this setting. This isolates the impact of semantic features, demonstrating their substantial capacity to modulate model behavior.
Full experiments, including EPR and TASS, along segmentation evaluation are provided in the Appendix.

\vspace{-2mm}
\section{Conclusion}
\label{sec:conclusion}

We presented C$^3$L, a framework that breaks the self-referential drift loop and mitigates explanation drift by anchoring representations to semantically grounded concepts. By regularizing attention toward semantic attributes rather than unstable pixel-level saliency, \method promotes learning of transferable features that better resist spurious correlations and explanation drifts. Our automated framework eliminates the need for manual annotation, making concept-based supervision scalable for continual learning. Extensive evaluation across five benchmarks demonstrates that \method establishes state-of-the-art performance in continual learning, particularly under challenging out-of-distribution shifts. Limitations and future work are detailed in the Appendix.

\begin{ack}
This work was supported by the Ministry of Innovation, Science \& Technology, Israel, and by the Israeli Science Foundation (ISF grant 977/26).
\end{ack}

{
    \small
    \bibliographystyle{unsrtnat}
    \bibliography{main}
}

\newpage
\appendix

\section{Appendix Overview}

This appendix provides comprehensive supplementary materials to support the main paper. We organize the content as follows:

\begin{itemize}
    \item \textbf{Appendix~\ref{sec:app_implementation}}: Detailed implementation specifications, hyperparameters, and training protocols.
    \item \textbf{Appendix~\ref{sec:app_design_just}}: Theoretical and empirical justification for method design choices.
    \item
    \textbf{Appendix~\ref{sec:app_crowding}}: Discussion and analysis about the crowding-out phenomenon.
    \item \textbf{Appendix~\ref{sec:app_additional_abl}}: Extended ablation studies including replay contribution, explanation methods, task granularity, and segmentation approaches.
    \item \textbf{Appendix~\ref{sec:app_experiments}}: Additional experiments on non-pretrained networks and results for Table~\ref{tab:results_cnn}.
    \item \textbf{Appendix~\ref{sec:app_xai_qual}}: Detailed explanation quality and robustness (RQ4) experimental details with full results.
    \item \textbf{Appendix~\ref{sec:app_foundation}}: Discussion on foundation model usage and fairness and PartImageNet experiment.
    \item \textbf{Appendix~\ref{sec:app_qualitative}}: Qualitative visualizations and case studies.
    \item \textbf{Appendix~\ref{sec:app_failure}}: Failure cases analysis.
    \item \textbf{Appendix~\ref{sec:app_limitations}}: Discussion on limitations of our method and future work.
\end{itemize}

\section{Implementation Details}
\label{sec:app_implementation}

\textbf{Full Implementation Details.} We use \textbf{ResNet-18}~\citep{he2016deep} (ImageNet-1K pretrained) for CNN experiments and ViT-B/16 \textbf{(ViT-B)}~\citep{dosovitskiy2020image} (ImageNet-21K pretrained) for transformer experiments. For explanation generation, we employ LRP~\citep{bach2015pixel} with $\epsilon$-rule ($\epsilon = 10^{-6}$) for CNNs and AttnLRP~\citep{achtibat2024attnlrp} for ViTs. For saliency map evaluation (e.g., Tab.~\ref{tab:segmentation}), explanations for all baselines are generated post-training using Grad-CAM~\citep{selvaraju2017grad} for CNNs and Grad-CAM-ViT~\citep{chefer2021transformer} for ViTs. While \method yields marginally higher scores when evaluated on its native LRP maps, we report Grad-CAM results to maintain a unified evaluation protocol. C$^3$L's saliency maps are min-max normalized to $[0,1]$ prior to BCE computation. Concept discovery follows the approach from~\citet{oikarinen2023label} which resulting in a concept set per class, filtered by occurrence rate ($\nu_{\text{occ}} \geq 0.15$) and coverage ($ \nu_{\text{cov}} \geq 0.2$). Spatial grounding uses SAM 3~\citep{carion2025sam} with confidence threshold $\tau_{\text{conf}} = 0.3$. We validate these configuration choices in Appendix~\ref{sec:app_additional_abl}.

\textbf{Full Training Details.} We apply regularization only to replay samples, as current task samples lack reference explanations. The replay buffer $\mathcal{M}$ (size $|\mathcal{M}| = 1000$) is populated via uniform random sampling from previous tasks.  Hyperparameters: $\lambda_{\text{concept}} = 1.0$, $\lambda_{\text{stab}} = 1.5$ for CNNs, and $\lambda_{\text{concept}} = 0.8$, $\lambda_{\text{stab}} = 1.2$ for ViTs. These values were selected based on validation set performance (validated in the Appendix). Results are averaged over 5 runs with different random seeds and task orderings. We report mean $\pm$ standard deviation and assess statistical significance ($p < 0.05$) using paired t-tests comparing \method against the best-performing baseline in each experimental setting.

\paragraph{Training Protocols.}
For our method, we employ stochastic gradient descent (SGD) for both model architectures, using a batch size of 128 and a learning rate of 0.001. Training is conducted for 20 epochs on ImageNet-R, ImageNet-A, ObjectNet, and DomainNet, and for 50 epochs on the CUB dataset.
All baselines were used according to the recommended setup reported in their papers and code.

\paragraph{Task Splits.} All datasets are split into tasks with an equal number of classes per task: CUB (200 classes), ImageNet-R (200), and ImageNet-A (200) into 10 tasks of 20 classes, and DomainNet (345 classes) into 5 tasks of 69 classes. ObjectNet contains 113 classes that overlap with ImageNet-1K, which cannot be divided evenly into 10 tasks; we therefore randomly exclude 3 classes and split the remaining 110 classes into 10 tasks of 11 classes each. All methods use the same class subset and splits.

\paragraph{Concept Discovery.}
We used the implementation from~\cite{oikarinen2023label} with GPT-4o-mini~\cite{hurst2024gpt} and used the prompts from the original github repository~\footnote{https://github.com/Trustworthy-ML-Lab/Label-free-CBM} without the "around" prompts.

\paragraph{Spatial Grounding Parameters.} For SAM 3 segmentation, we use confidence threshold $\tau_{\text{conf}} = 0.3$ and used the Ultralytics implementation~\footnote{https://docs.ultralytics.com/models/sam-3/}. Lower thresholds increase false positives (irrelevant regions), while higher thresholds miss valid concept instances. We validate this choice in Appendix~\ref{app:sam3_threshold}.

\paragraph{Replay Buffer.} The replay buffer size is set to $|\mathcal{M}| = 1000$ for fair comparison with baselines. We use uniform random sampling across previous tasks to populate the buffer. Each task contributes equally to maintain balanced representation.

\paragraph{Computational Environment.} All experiments are conducted on NVIDIA A100 GPUs using PyTorch. For both ResNet-18 and ViT-B experiments, we use single-GPU training.

\section{Method Design Justification}
\label{sec:app_design_just}

\subsection{Background Regularization Strategy}
\label{sec:background_reg}
A central design choice in \method is how to regularize non-concept (background) regions. Unlike pixel-level alignment methods that indiscriminately anchor explanations to a reference saliency map, \method introduces a semantic prior that explicitly reallocates relevance toward intrinsic class attributes while constraining background behavior.

\paragraph{Rationale.}
Our design is motivated by three considerations.

First, \textit{mitigating spurious correlations}: by enforcing full relevance within discovered concept regions, we can leverage the finite nature of the relevance budget. While relevance conservation is approximate in modern deep networks~\citep{montavon2019layer}, empirically, maximizing relevance on foreground concepts  encourages a redistribution of mass away from the background. This ``crowding-out'' effect helps suppress spurious cues without requiring explicit penalization.
Second, \textit{incomplete concept discovery}: automated concept generation may omit valid diagnostic features (e.g., fine-grained parts). Explicit background suppression risks eliminating such implicitly learned cues present in the pretrained model.
Third, \textit{context preservation}: background regions encode stable contextual information (e.g., pose, geometry, scale) learned during pretraining, which can support generalization and should not be erased.

To balance these factors, $\mathcal{L}_{\text{stab}}$ enforces \emph{stability}, preserving the spatial structure of background relevance, rather than suppression. While the relative magnitude of background relevance naturally decreases due to relevance reallocation, anchoring its pattern prevents drift toward new, task-specific artifacts.

\paragraph{Empirical Support.}
We compare three strategies: (i) the full \method, (ii) no background regularization (w/o $\mathcal{L}_{\text{stab}}$ on Section~\ref{sec:ablations}), and (iii) explicit background suppression without replay (\method-NoReplay on Section~\ref{sec:abl_replay_analysis}). Experiments with ViT-B on ImageNet-R and CUB (Table~\ref{tab:abl_background}) show that stabilization outperforms both alternatives; since (iii) also removes replay, the comparison with (ii) is the one that isolates the background term, confirming that preserving pretrained contextual priors while preventing spurious correlations yields the best trade-off.

\begin{table}[t]
\centering
\caption{Evaluation on different background design for \method using ViT-B on ImageNet-R and CUB.}
\label{tab:abl_background}
\resizebox{0.65\columnwidth}{!}{
\begin{tabular}{l|ccc|ccc}
\toprule
\multirow{2}{*}{{Configuration}} & \multicolumn{3}{c|}{{ImageNet-R}} & \multicolumn{3}{c}{{CUB}} \\
 & ACC $\uparrow$ & F $\downarrow$ & AIA $\uparrow$ & ACC $\uparrow$ & F $\downarrow$ & AIA $\uparrow$ \\
\midrule
\method & 82.74 & 1.89 & 83.92 & 83.65 & 2.92 & 84.64 \\
\method-NoReplay  & 79.82 & 3.84  & 78.99 & 80.94 & 3.27  & 81.78 \\
w/o $\mathcal{L}_{\text{stab}}$ & 78.94 & 3.01 & 78.32 & 78.42 & 3.86 & 78.98 \\
\bottomrule
\end{tabular}
}
\end{table}

\section{Quantitative Evidence for the Crowding-Out Effect}
\label{sec:app_crowding}
In Section~\ref{sec:cccl}, we introduced the concept of \emph{relevance reallocation} as a mechanism by which enforcing high attribution within concept regions indirectly suppresses spurious background correlations. This phenomenon, which we term the \emph{crowding-out effect}, operates through the approximate conservation of relevance mass in neural networks~\citep{montavon2019layer}. Specifically, when the model is encouraged to concentrate relevance within semantically meaningful concept regions via $\mathcal{L}_{\text{concept}}$, the finite relevance budget naturally reallocates away from non-diagnostic background regions, even without explicit suppression.

While Section~\ref{sec:background_reg} provided empirical evidence via ablation studies, a critical question remains: does relevance mass actually move from background to concept regions during training, or do both regions simply change independently? To directly test the crowding-out hypothesis, we track the spatial distribution of relevance across different image regions throughout continual learning.

\subsection{Relevance Mass Flow Analysis}

\paragraph{Experimental Protocol.} We train \method on CUB using the standard 10-task split (20 classes per task) with a ViT-B backbone. For each sample $x$ in the test set of task $\mathcal{T}_1$, we partition pixels into three regions: \\(1) \textbf{Concept regions (C):} Pixels within $M_{\text{concept}}^{(x)}$, corresponding to discovered semantic attributes.
    \\(2) \textbf{Background regions (B):} Pixels within $M_{\text{bg}}^{(x)}$, representing non-concept areas.
    \\(3) \textbf{Total (T):} All pixels in the image.

At the end of each task $\mathcal{T} \in \{1, 3, 5, 8, 10\}$, we freeze the model and compute relevance maps $S_x$ using AttnLRP~\citep{achtibat2024attnlrp}. For each region $R \in \{C, B, T\}$, we measure:

\begin{equation}
R_{\text{total}}(R) = \sum_{i \in R} S_x(i), \quad R_{\text{rel}}(R) = \frac{R_{\text{total}}(R)}{f_\theta(x)_y}
\end{equation}

where $R_{\text{total}}(R)$ is the (signed) relevance assigned to region $R$ and $R_{\text{rel}}(R)$ is the fraction of the target logit $f_\theta(x)_y$ attributed to it. Under approximate conservation (Eq.~\ref{eq:conservation}), $R_{\text{rel}}(T) \approx 1$ and $R_{\text{rel}}(C) \approx \rho_{\mathcal{C}}$, the concept share defined in Section~\ref{sec:cccl}. We also compute the \textbf{Concentration Ratio (CR)}:

\begin{equation}
\text{CR} = \frac{R_{\text{rel}}(C)}{R_{\text{rel}}(B)}
\end{equation}

which directly captures the shift of relevance from background to concept regions. Under the crowding-out hypothesis, we expect:
(1) $R_{\text{rel}}(C)$ to increase monotonically with training,
    (2) $R_{\text{rel}}(B)$ to decrease monotonically,
    (3) CR to increase, reflecting concentration toward concepts,
    (4) $R_{\text{rel}}(C) + R_{\text{rel}}(B) = R_{\text{rel}}(T)$ to remain close to 1 (approximate relevance conservation).

We average these metrics over all test samples from $\mathcal{T}_1$ and report mean values.

\subsection{Results and Analysis}

Table~\ref{tab:crowding_out} presents the relevance mass flow analysis. The results provide strong quantitative evidence for the crowding-out effect:

\begin{table}[t]
\centering
\caption{Relevance mass flow analysis across training tasks on CUB $\mathcal{T}_1$ test set. Results demonstrate reallocation of relevance from background to concept regions, validating the crowding-out mechanism.}
\label{tab:crowding_out}
\begin{tabular}{@{}lcccc@{}}
\toprule
\textbf{Training Stage} & $R_{\text{rel}}(C)$ $\uparrow$ & $R_{\text{rel}}(B)$ $\downarrow$ & {CR} $\uparrow$ & $R_{\text{rel}}(T)$ \\
\midrule
After $\mathcal{T}_1$   & 0.52 & 0.46 & 1.13 & 0.98 \\
After $\mathcal{T}_3$   & 0.61 & 0.37 & 1.65 & 0.98 \\
After $\mathcal{T}_5$   & 0.65 & 0.33 & 1.97 & 0.98 \\
After $\mathcal{T}_8$   & 0.71 & 0.28 & 2.54 & 0.99 \\
After $\mathcal{T}_{10}$ & 0.72 & 0.28 & 2.57 & 1.00 \\
\midrule
Absolute Change & +0.20 & -0.18 & +1.44 & +0.02 \\
Relative Change & +38.5\% & -39.1\% & +127.4\% & -- \\
\bottomrule
\end{tabular}
\end{table}

As training progresses from $\mathcal{T}_1$ to $\mathcal{T}_{10}$, the fraction of the logit attributed to concept regions increases, while background relevance decreases. This shift occurs despite the model learning 180 new classes across subsequent tasks, which typically induces explanation drift toward spurious features~\citep{ebrahimi2020remembering}. The concentration ratio CR increases from 1.13 after $\mathcal{T}_1$ to 2.57 after $\mathcal{T}_{10}$. This demonstrates that relevance becomes progressively more concentrated in semantically meaningful concept regions relative to backgrounds.
The relevance maps account for 98--100\% of the target logit at every stage ($R_{\text{rel}}(T) \in [0.98, 1.00]$), so approximate conservation holds throughout training. The 0.20 gain in concept share is therefore matched by a 0.18 reduction in background share: relevance is reallocated from background to concept regions rather than added on top of it, consistent with the crowding-out effect described in Section~\ref{sec:cccl}.

These results validate the theoretical mechanism proposed in Section~\ref{sec:cccl}: by enforcing high relevance within concept regions through $\mathcal{L}_{\text{concept}}$, the model satisfies this constraint by reallocating relevance mass away from background regions. The background stabilization term $\mathcal{L}_{\text{stab}}$ prevents drift to new spurious artifacts while allowing this magnitude reduction. Together, the dual regularization creates a \emph{crowding-out effect} where semantically grounded concepts compete for and capture an increasing share of the relevance budget, naturally suppressing background correlations without explicit penalization.

This analysis provides the first direct quantitative evidence for the crowding-out effect in concept-based continual learning. Unlike previous work that measures only final explanation quality~\citep{ebrahimi2020remembering,saha2023saliency}, tracking relevance flow throughout training reveals the mechanism by which concept regularization mitigates catastrophic forgetting. The reallocation of relevance from backgrounds to concepts demonstrates that semantic anchoring actively reshapes the model's decision-making process, not merely preserving existing attention patterns.

Importantly, this redistribution occurs without explicit background suppression. The crowding-out effect emerges naturally from the interaction between concept overlap enforcement ($\mathcal{L}_{\text{concept}}$), background stability ($\mathcal{L}_{\text{stab}}$), and approximate relevance conservation. This validates our design choice in favor of stability over suppression (Appendix~\ref{sec:background_reg}): while background relevance \emph{magnitude} decreases due to crowding-out, its \emph{spatial pattern} remains anchored to pretrained priors, preserving useful contextual information while preventing spurious correlations.

\section{Extended Ablation Studies}
\label{sec:app_additional_abl}

\subsection{Impact of Replay in Semantic-Centered Training}
\label{sec:abl_replay_analysis}

To disentangle the respective contributions of experience replay and semantic regularization, we evaluate \method under four distinct configurations:

\begin{enumerate}
    \item \textbf{\method}: our full method.

    \item \textbf{\method-NoReplay}: a replay-free variant that performs class-incremental training using only semantic regularization on the current task data. In this setting, no replay buffer is maintained and training is conducted solely on samples from the newly introduced task, with saliency maps generated on-the-fly during optimization. Concept discovery is performed once prior to training each task, while concept grounding is applied at the sample level during training. The objective is to explicitly align the model's explanations with concept-specific regions, using the following regularization terms in addition to the task loss $\mathcal{L}_{\text{task}}$:

        \begin{equation}
        \mathcal{L}_{\text{concept}} =
        \text{BCE}( S_x \odot M_{\text{concept}}^{(x)} , M_{\text{concept}}^{(x)})
        \label{eq:concept_loss2}
        \end{equation}

        \begin{equation}
        \mathcal{L}_{\text{stab}} =
        \left\| M_{\text{bg}}^{(x)} \odot S_x \right\|_2^2
        \label{eq:stab_loss2}
        \end{equation}

    \item \textbf{\method-FullReplay}: a variant that applies experience replay to both semantic regularization terms. This configuration follows class-incremental training with concept regularization on both current and replayed samples, using the same replay mechanism as \method and augmenting $\mathcal{L}_{\text{task}}$ with the following losses:

\begin{equation}
\mathcal{L}_{\text{concept}} = \frac{1}{|\mathcal{M}|} \sum_{(x,y) \in \mathcal{M}}
\text{BCE}( M_{\text{concept}}^{(x)} \odot  S_x , M_{\text{concept}}^{(x)} \odot S_x^{\text{ref}} )
\label{eq:concept_loss3}
\end{equation}

\begin{equation}
\mathcal{L}_{\text{stab}} = \frac{1}{|\mathcal{M}|} \sum_{(x,y) \in \mathcal{M}}
\left\| M_{\text{bg}}^{(x)} \odot \left( S_x - S_x^{\text{ref}} \right) \right\|_2^2
\label{eq:stab_loss3}
\end{equation}

    \item \textbf{ER}: standard experience replay without any semantic or explanation-based regularization.
\end{enumerate}

Table~\ref{tab:replay_analysis} reports results on ImageNet-R and CUB using a ViT-B backbone. The full \method consistently achieves the best overall performance, indicating that combining positive supervision on concept regions with stabilizing constraints on non-concept areas yields the most effective trade-off between plasticity and stability.
Essentially, \method-FullReplay is very similar to RRR-ViT-AttnLRP, as it preserves explanation consistency while applying a slightly different distance function.

Notably, \method-NoReplay outperforms standard ER despite not storing any past samples. Moreover, it surpasses \method-FullReplay on the CUB dataset and achieves comparable performance on ImageNet-R. These findings suggest that semantic regularization alone can serve as an alternative to experience replay, particularly in scenarios where storing training data is infeasible due to privacy, memory, or deployment constraints. Intuitively, anchoring learning to semantically meaningful concepts identified during pretraining encourages the model to preserve its reasoning patterns across tasks, even in the absence of explicit replayed examples.

\begin{table}[t]
\centering

\caption{Ablation study disentangling the contribution of experience replay versus semantic regularization on ImageNet-R and CUB (ViT-B).}
\label{tab:replay_analysis}
\scalebox{0.8}{
\begin{tabular}{lccccccc}
\toprule
& \multicolumn{3}{c}{\textbf{ImageNet-R}} & & \multicolumn{3}{c}{\textbf{CUB}} \\
\cmidrule{2-4} \cmidrule{6-8}
\textbf{Method} & ACC $\uparrow$ & F $\downarrow$ & AIA $\uparrow$ & & ACC $\uparrow$ & F $\downarrow$ & AIA $\uparrow$ \\
\midrule
ER & 72.27${\pm 0.82}$ & 9.13${\pm 0.54}$ & 71.56${\pm 0.71}$ & & 70.92${\pm 0.94}$ & 8.48${\pm 0.62}$ & 71.40${\pm 0.88}$ \\
\method-NoReplay  & 79.82${\pm 0.65}$ & 3.84${\pm 0.42}$  & 78.99${\pm 0.58}$ & & 80.94${\pm 0.72}$ & 3.27${\pm 0.38}$  & 81.78${\pm 0.66}$ \\
\method-FullReplay & 80.32${\pm 0.58}$ & 2.96${\pm 0.39}$  & 80.04${\pm 0.52}$ & & 80.13${\pm 0.64}$ & 3.42${\pm 0.51}$  & 81.52${\pm 0.59}$ \\
\method      & $\mathbf{82.74{\pm 0.76}}$ & $\mathbf{1.89{\pm 0.33}}$ & $\mathbf{83.92{\pm 0.45}}$ & & $\mathbf{83.65{\pm 0.28}}$ & $\mathbf{2.92{\pm 0.24}}$ & $\mathbf{84.64{\pm 0.37}}$ \\
\bottomrule
\end{tabular}}
\end{table}

\subsection{Explanation method}
\label{sec:abl_xai}
\begin{table}[h]
\centering

\caption{Performance comparison using different explanation methods (AttnLRP, Grad-CAM, IIA, T-Attr) within the \method framework (ViT-B, ImageNet-R).}
\label{tab:abl_xai}
\begin{tabular}{lccc}
\toprule
\textbf{Explanation Method} & \textbf{ACC $\uparrow$} & \textbf{F $\downarrow$} & \textbf{AIA $\uparrow$} \\
\midrule
\textbf{\method-AttnLRP (Ours)} & $82.74 \pm 0.76$ & $1.89 \pm 0.33$ & $83.92 \pm 0.45$ \\
\method-GradCAM & $80.92 \pm 0.84$ & $2.43 \pm 0.41$ & $82.15 \pm 0.52$ \\
\method-IIA & $81.85 \pm 0.77$ & $2.08 \pm 0.35$ & $83.12 \pm 0.46$ \\
\method-T-Attr & $81.47 \pm 0.79$ & $2.21 \pm 0.37$ & $82.68 \pm 0.48$ \\
\bottomrule
\end{tabular}
\end{table}

We compare AttnLRP against three alternatives: Grad-CAM~\citep{selvaraju2017grad}, Integrated Iterated Attribution (IIA)~\citep{barkan2023visual}, and Transformer Attribution (T-Attr)~\citep{chefer2021transformer}. To conduct this analysis, we implemented three additional \method variants, each replacing the LRP-based explanation creation with the corresponding baseline method. Table~\ref{tab:abl_xai} presents the results (method names reflect their adjusted \method variants). The findings show a clear advantage for LRP in this process, likely because LRP propagates relevance through multiple layers, offering a more comprehensive view of the model’s internal representations. However, the relatively small gap indicates that \method is somewhat robust to explanation method choice, as long as the method produces reasonably faithful attributions.

\subsection{Impact of Task Granularity}
\label{sec:task_splitting}

To evaluate the robustness of $C^3L$ under varying task granularities, we conduct an ablation study on ImageNet-R by varying the number of splits. Following the protocol in~\cite{smith2023coda}, we divide the 200 classes of ImageNet-R into $T \in \{5, 10, 20\}$ tasks, corresponding to 40, 20, and 10 classes per task, respectively. We compare our method against the top-performing replay-augmented baselines from Table~\ref{tab:results_cnn} using the ViT-B backbone.

As presented in Table~\ref{tab:task_split_ablation}, performance naturally degrades for all methods as the number of tasks increases, due to the higher frequency of model updates and the resulting accumulation of catastrophic forgetting steps. However, $C^3L$ consistently maintains the highest accuracy performance (ACC and AIA) and lowest forgetting (F) across all settings. Notably, in the challenging 20-task setting, where forgetting is typically most severe, \method outperforms the nearest competitor (PLAN-R) by a significant margin. This demonstrates that anchoring representations to stable semantic concepts provides a more durable protection against forgetting than architectural or prompt-based regularization alone, particularly in long-sequence scenarios.

\begin{table}[h]
\centering

\caption{Robustness analysis under varying task granularities (5, 10, and 20 split tasks) on ImageNet-R using the ViT-B backbone.}
\label{tab:task_split_ablation}
\resizebox{0.65\columnwidth}{!}{%
\setlength{\tabcolsep}{2pt}
\begin{tabular}{l ccc ccc ccc}
\toprule
\multirow{2}{*}{{Method}} & \multicolumn{3}{c}{{5 Tasks}} & \multicolumn{3}{c}{{10 Tasks}} & \multicolumn{3}{c}{{20 Tasks}} \\
\cmidrule(lr){2-4} \cmidrule(lr){5-7} \cmidrule(lr){8-10}
& {ACC} $\uparrow$ & {F} $\downarrow$ & {AIA} $\uparrow$ & {ACC} $\uparrow$ & {F} $\downarrow$ & {AIA} $\uparrow$ & {ACC} $\uparrow$ & {F} $\downarrow$ & {AIA} $\uparrow$ \\
\midrule
L2P-R      & 80.52 & 3.15 & 82.13 & 76.82 & 3.91 & 77.34 & 74.12 & 7.85 & 75.20 \\
CODA-R        & 81.45 & 2.68 & 82.90 & 78.54 & 2.82 & 79.52 & 75.30 & 6.12 & 76.85 \\
MOS-R           & 82.37 & 1.95 & 83.86 & 80.92 & 2.43 & 81.55 & 76.45 & 4.88 & 78.10 \\
PLAN-R         & 82.18 & 2.09 & 83.49 & 80.13 & 2.58 & 80.89 & 76.90 & 4.50 & 78.55 \\
\midrule
\textbf{C\textsuperscript{3}L} & \textbf{84.63} & \textbf{1.10} & \textbf{86.15} & \textbf{82.74} & \textbf{1.89} & \textbf{83.92} & \textbf{81.43} & \textbf{2.21} & \textbf{82.38} \\
\bottomrule
\end{tabular}
}
\end{table}

\subsection{Concept-Level vs. Object-Level Segmentation}
\label{sec:abl_concept_source}
Table~\ref{tab:concept_vs_object} presents results for ViT-B on the ImageNet-R and CUB datasets.
To assess whether fine-grained concept supervision provides advantages over coarse object-level segmentation, we replace the discovered concept masks with object segmentation masks. Specifically, we prompt SAM~3 using class names (e.g., ``siberian husky'') instead of concept lists (e.g., ``blue eyes, fluffy ears, pointed snout''). This substitution leads to a consistent degradation in performance, indicating that concept-level supervision captures sub-object parts and attributes (e.g., ``curved beak'' or ``striped wing'') that offer more precise and informative guidance than binary object masks.

While object segmentation conveys category-level information (i.e., identifying \emph{what} object is present), it does not specify \emph{which} visual features are diagnostically relevant for classification. In contrast, concept-level masks emphasize discriminative attributes, resulting in denser and more localized supervision that more effectively constrains model attention. Additionally, object-level segmentations often encompass redundant or weakly informative regions, for example, assigning high importance to the entire body of a ``Brown Bear'', whereas concept-level supervision can focus on a minimal yet sufficient subset of salient parts.

This experiment isolates the contribution of concept-level guidance. Although ``\method (Object segments)'' relies on supervision from the same foundation model as \method (SAM 3), its performance is substantially inferior, indicating that the gains stem from semantic concept guidance rather than from the foundation model itself.

\begin{table}[h]
\centering
\caption{Performance comparison between using fine-grained concept masks versus coarse object-level segmentation masks (ViT-B).}
\label{tab:concept_vs_object}
\resizebox{\textwidth}{!}{%
\begin{tabular}{lcccccc}
\toprule
& \multicolumn{3}{c}{{ImageNet-R}} & \multicolumn{3}{c}{{CUB}} \\
\cmidrule(lr){2-4} \cmidrule(lr){5-7}
{Mask Type} & {ACC $\uparrow$} & {F $\downarrow$} & {AIA $\uparrow$} & {ACC $\uparrow$} & {F $\downarrow$} & {AIA $\uparrow$} \\
\midrule
\method & $82.74 \pm 0.76$ & $1.89 \pm 0.33$ & $83.92 \pm 0.45$ & $83.65 \pm 0.28$ & $2.92 \pm 0.24$ & $84.64 \pm 0.37$ \\
\method (Object segments) & $79.93 \pm 0.91$ & $2.84 \pm 0.41$ & $81.06 \pm 0.58$ & $78.47 \pm 1.24$ & $4.01 \pm 0.66$ & $81.64 \pm 0.38$ \\
\bottomrule
\end{tabular}%
}
\end{table}

\subsection{SAM 3 vs. Grounded SAM}
\label{sec:abl_spatial_grounding}
\begin{table}[h]
\centering
\caption{Impact of the spatial grounding foundation model: comparison between SAM 3 and Grounded SAM on ImageNet-R (ViT-B).}
\label{tab:grounding_comparison}
\begin{tabular}{lccc}
\toprule
{Grounding Method} & {ACC $\uparrow$} & {F $\downarrow$} & {AIA $\uparrow$} \\
\midrule
C$^3$L + Grounded SAM & $81.98 \pm 0.82$ & $2.12 \pm 0.38$ & $83.25 \pm 0.73$ \\
C$^3$L & $\mathbf{82.74 \pm 0.76}$ & $\mathbf{1.89 \pm 0.33}$ & $\mathbf{83.92 \pm 0.45}$ \\
\bottomrule
\end{tabular}
\end{table}

\paragraph{Prompt Complexity Robustness Analysis.}
Prior work suggests that GroundedSAM underperforms on complex or verbose prompts~\cite{van2024guided,zeng_exploration_2024}. To further justify the use of SAM~3 over GroundedSAM (Table~\ref{tab:grounding_comparison}), we conduct a targeted evaluation of robustness to prompt complexity. We randomly sample 50 classes from the CUB dataset and filter their automatically discovered concept sets to include concepts more likely to be  verbose, fine-grained descriptions (i.e., concepts strictly longer than two words), resulting in a set of 286 textual concepts. We evaluate these concepts across 500 images (10 randomly selected images per class), prompting both SAM~3 and GroundedSAM with the exact same text prompts. We measure the detection hit-rate, defined as the percentage of prompts yielding a segmentation mask with a confidence score $\ge 0.3$, alongside the average confidence score across all predicted masks. SAM 3 maintains a high hit-rate of 86\% with an average confidence of 0.67, whereas GroundedSAM's hit-rate drops significantly to 59\% with an average confidence of 0.45. These findings suggest that GroundedSAM's ability to localize fine-grained, verbose semantic attributes is inferior to that of SAM~3.
Notably, this establishes the C$^{3}$L + GroundedSAM vs. C$^{3}$L ablation (Table~\ref{tab:grounding_comparison}) not merely as a foundation model comparison, but as a noise robustness evaluation for our framework. Despite the degradation in mask quality and the omission rate introduced by GroundedSAM, C$^{3}$L retains competitive continual learning performance, underscoring the resilience of our dual-regularization objective to noisy or incomplete semantic grounding.

\paragraph{Results.}
As we can see on Table~\ref{tab:grounding_comparison} replacing SAM 3 with GroundedSAM~\citep{ren2024grounded} for concept localization slightly decreases performance. SAM 3's advantage lies in its ability to handle multiple concepts in a single batched query (one noun-phrase prompt per concept), returning confidence-scored masks that enable more accurate aggregation. GroundedSAM processes concepts sequentially, requiring multiple forward passes. Nevertheless, the modest performance gap indicates that \method is relatively robust to grounding method choice - as long as concepts are reasonably well-localized, the framework provides substantial benefits. Finally, GroundedSAM is more suitable for object segmentation rather than concept segmentation~\cite{carion2025sam}.

\subsection{Concept Filtering: Required vs. Optional}
\label{app:validation_ablation}

We evaluate whether the concept filtering step (\S\ref{sec:concept_discovery_grounding}) is necessary by comparing performance with and without filtering.

\begin{table}[h]
\centering

\caption{Ablation study on the necessity of the concept filtering and validation step across ImageNet-R and CUB.}
\label{tab:validation_ablation}
\scalebox{0.75}{
\begin{tabular}{lcccccc}
\toprule
& \multicolumn{3}{c}{{ImageNet-R}} & \multicolumn{3}{c}{{CUB}} \\
\cmidrule(lr){2-4} \cmidrule(lr){5-7}
{Configuration} & {ACC $\uparrow$} & {F $\downarrow$} & {AIA $\uparrow$} & {ACC $\uparrow$} & {F $\downarrow$} & {AIA $\uparrow$} \\
\midrule
C$^3$L (no validation) & $81.93 \pm 0.84$ & $2.17 \pm 0.38$ & $83.15 \pm 0.52$ & $82.47 \pm 0.89$ & $3.24 \pm 0.41$ & $83.52 \pm 0.76$ \\
C$^3$L (with validation) & $\mathbf{82.74 \pm 0.76}$ & $\mathbf{1.89 \pm 0.33}$ & $\mathbf{83.92 \pm 0.45}$ & $\mathbf{83.65 \pm 0.28}$ & $\mathbf{2.92 \pm 0.24}$ & $\mathbf{84.64 \pm 0.37}$ \\
\midrule
MOS-R (Best baseline) & {80.92${\pm 0.42}$} & {2.43${\pm 0.28}$} & {81.55${\pm 0.31}$} & 78.82${\pm 0.65}$ & 3.84${\pm 0.29}$ & {82.52${\pm 0.53}$}\\
\bottomrule
\end{tabular}}
\end{table}

Filtering provides modest improvements, but even unvalidated concepts substantially outperform the best baseline. This demonstrates robustness: GPT-4o-mini generates predominantly valid concepts, and the framework tolerates some noisy concepts through the stability term's robustness.

\subsection{SAM 3 Confidence Threshold}
\label{app:sam3_threshold}

We evaluate sensitivity to the SAM 3 confidence threshold $\tau_{\text{conf}} \in \{0.2, 0.3, 0.4, 0.5\}$. Table~\ref{tab:threshold_ablation} present these results.

\begin{table}[h]
\centering
\caption{Sensitivity analysis of the SAM 3 confidence threshold ($\tau_{conf}$) for concept grounding on ImageNet-R (ViT-B).}
\label{tab:threshold_ablation}
\begin{tabular}{lccc}
\toprule
$\tau_{\text{conf}}$ & {ACC $\uparrow$} & {F $\downarrow$} & {AIA $\uparrow$} \\
\midrule
0.2 & $82.15 \pm 0.85$ & $2.08 \pm 0.61$ & $83.24 \pm 0.62$ \\
0.3 (our choice) & $\mathbf{82.74 \pm 0.76}$ & $\mathbf{1.89 \pm 0.33}$ & $\mathbf{83.92 \pm 0.45}$ \\
0.4 & $82.43 \pm 0.79$ & $1.95 \pm 0.46$ & $83.61 \pm 0.48$ \\
0.5 & $81.87 \pm 0.83$ & $2.14 \pm 0.79$ & $82.98 \pm 0.95$ \\
\bottomrule
\end{tabular}
\end{table}

We can observe that $\tau_{\text{conf}} = 0.3$ yields the best overall performance. Lower thresholds introduce noisy false-positive regions, whereas higher thresholds exclude valid concept instances. Notably, performance varies smoothly across a wide range of threshold values, indicating that the method is not overly sensitive to the specific choice of $\tau_{\text{conf}}$.

\subsection{Hyperparameter Sensitivity Analysis}
\label{sec:abl_hyperparams}

Table~\ref{tab:hyperparam_sensitivity} shows performance across different values of $\lambda_{\text{concept}}$ and $\lambda_{\text{stab}}$ on ImageNet-R (ViT-B).

\begin{table}[h]
\centering
\caption{Sensitivity analysis for regularization hyperparameters $\lambda_{concept}$ and $\lambda_{stab}$ on ImageNet-R (ViT-B).}
\label{tab:hyperparam_sensitivity}
\begin{tabular}{ccccc}
\toprule
$\lambda_{\text{concept}}$ & $\lambda_{\text{stab}}$ & {ACC $\uparrow$} & {F $\downarrow$} & {AIA $\uparrow$} \\
\midrule
0.5 & 0.6 & $80.83 \pm 0.89$ & $2.34 \pm 0.41$ & $81.34 \pm 0.64$ \\
\textbf{0.8} & \textbf{1.2} & $82.74 \pm 0.76$ & $1.89 \pm 0.33$ & $83.92 \pm 0.45$ \\
1.0 & 1.5 & $82.51 \pm 0.81$ & $1.95 \pm 0.36$ & $82.42 \pm 0.81$ \\
1.5 & 2.0 & $81.92 \pm 0.87$ & $2.11 \pm 0.38$ & $81.85 \pm 0.75$ \\
2.0 & 2.5 & $81.24 \pm 0.93$ & $2.28 \pm 0.42$ & $81.37 \pm 0.53$ \\
\bottomrule
\end{tabular}
\end{table}

Performance only slightly varies across the tested range, indicating reasonable robustness to hyperparameter choices. We select $\lambda_{\text{concept}}=0.8$, $\lambda_{\text{stab}}=1.2$ based on validation set performance.

\section{Additional Experiments}
\label{sec:app_experiments}

\subsection{Performance on Non-Pretrained Networks}

To assess whether C$^3$L's benefits extend beyond pretrained networks, we evaluate on ResNet-18 trained from random initialization on CUB.

\begin{table}[h]
\centering
\caption{Continual learning performance on CUB using a non-pretrained ResNet-18 backbone.}
\label{tab:non_pretrained}
\begin{tabular}{lccc}
\toprule
{Method} & {ACC $\uparrow$} & {F $\downarrow$} & {AIA $\uparrow$} \\
\midrule
ER & $42.17 \pm 1.35$ & $18.94 \pm 0.92$ & $45.83 \pm 1.24$ \\
DER++ & $48.23 \pm 1.18$ & $14.62 \pm 0.78$ & $51.47 \pm 1.09$ \\
EPR & $51.02 \pm 0.98$ & $12.84 \pm 0.26$ & $53.57 \pm 0.75$ \\
RRR & $49.84 \pm 1.24$ & $13.95 \pm 0.82$ & $52.93 \pm 1.15$ \\
C$^3$L & $\mathbf{59.21 \pm 1.08}$ & $\mathbf{10.74 \pm 0.71}$ & $\mathbf{60.18 \pm 0.97}$ \\
\midrule
Joint Training & $68.45 \pm 0.89$ & -- & -- \\
\bottomrule
\end{tabular}
\end{table}

\method outperforms baselines by large margin also without pretraining, though absolute performance is lower than with ImageNet pretraining.

 \subsection{Full Continual Learning Results}
Table~\ref{tab:results_cnn_full} presents full results for Table~\ref{tab:results_cnn} from the main paper. C$^3$L's with $\mathcal{L}_{\text{concept}}$ applied to all training samples, is denoted as \method{}++. Unlike \method, however, \method{}++ does not follow a continual learning replay setting and is therefore not considered our primary method. Since Joint Training uses no concept supervision, \method{}++ can exceed it (e.g., on CUB).
\begin{table*}[t]
\centering

\caption{Quantitative comparison of Continual Learning performance across five benchmarks using ResNet-18 and ViT-B architectures. We report Average Accuracy (ACC), Average Forgetting (F), and Average Incremental Accuracy (AIA). We report results over 5 trials.}

\label{tab:results_cnn_full}
\resizebox{\textwidth}{!}{
\begin{tabular}{l|ccc|ccc|ccc|ccc|ccc}
\toprule
& \multicolumn{3}{c|}{{CUB}} & \multicolumn{3}{c|}{{DomainNet}} & \multicolumn{3}{c|}{{ImageNet-R}} & \multicolumn{3}{c|}{{ImageNet-A}} & \multicolumn{3}{c}{{ObjectNet}} \\
\textbf{Method} & ACC $\uparrow$ & F $\downarrow$ & AIA $\uparrow$ & ACC $\uparrow$ & F $\downarrow$ & AIA $\uparrow$ & ACC $\uparrow$ & F $\downarrow$ & AIA $\uparrow$ & ACC $\uparrow$ & F $\downarrow$ & AIA $\uparrow$ & ACC $\uparrow$ & F $\downarrow$ & AIA $\uparrow$ \\
\midrule
\multicolumn{16}{c}{\emph{ResNet-18}}\\
\midrule
Joint Training & 83.04${\pm 0.51}$ & - & - & 74.99${\pm 0.22}$ & - & - & 76.31${\pm 0.90}$ & - & - & 71.22${\pm 0.42}$ & - & - & 73.42${\pm 0.91}$ & - & - \\
\midrule
\multicolumn{16}{l}{\emph{Regularization-based}} \\
EWC & 55.15${\pm 0.95}$ & 22.34${\pm 0.72}$ & 60.82${\pm 1.05}$ & 54.40${\pm 0.88}$ & 14.25${\pm 0.64}$ & 54.81${\pm 0.92}$ & 55.24${\pm 0.77}$ & 18.42${\pm 0.58}$ & 58.36${\pm 0.83}$ & 48.04${\pm 0.61}$ & 28.52${\pm 1.15}$ & 52.75${\pm 0.55}$ & 48.42${\pm 0.68}$ & 26.24${\pm 0.79}$ & 52.04${\pm 0.73}$ \\
LwF & 57.62${\pm 0.83}$ & 20.15${\pm 0.65}$ & 62.34${\pm 0.92}$ & 56.27${\pm 0.75}$ & 12.82${\pm 0.52}$ & 56.53${\pm 0.84}$ & 57.81${\pm 0.69}$ & 16.85${\pm 0.48}$ & 60.92${\pm 0.76}$ & 48.83${\pm 0.54}$ & 26.83${\pm 0.98}$ & 54.11${\pm 0.49}$ & 49.94${\pm 0.62}$ & 24.12${\pm 0.71}$ & 53.73${\pm 0.66}$ \\

PEC & 62.23${\pm 0.76}$ & 15.94${\pm 0.62}$ & 66.71${\pm 0.85}$ & 60.34${\pm 0.68}$ & 9.82${\pm 0.45}$ & 60.42${\pm 0.72}$ & 62.12${\pm 0.58}$ & 12.93${\pm 0.52}$ & 64.84${\pm 0.65}$ & 52.74${\pm 0.48}$ & 22.14${\pm 0.82}$ & 58.01${\pm 0.44}$ & 55.02${\pm 0.55}$ & 19.82${\pm 0.64}$ & 59.20${\pm 0.59}$ \\

\multicolumn{16}{l}{\emph{Replay-based}} \\
ER & 64.34${\pm 0.92}$ & 14.25${\pm 0.58}$ & 68.42${\pm 0.84}$ & 61.92${\pm 0.81}$ & 8.92${\pm 0.55}$ & 61.99${\pm 0.76}$ & 63.38${\pm 0.68}$ & 11.85${\pm 0.49}$ & 66.15${\pm 0.72}$ & 55.01${\pm 0.45}$ & 19.43${\pm 0.66}$ & 59.80${\pm 0.38}$ & 57.84${\pm 0.52}$ & 17.24${\pm 0.58}$ & 62.01${\pm 0.55}$ \\
iCaRL & 66.82${\pm 0.88}$ & 12.74${\pm 0.54}$ & 70.65${\pm 0.79}$ & 64.00${\pm 0.75}$ & 7.85${\pm 0.48}$ & 63.92${\pm 0.69}$ & 65.72${\pm 0.64}$ & 10.43${\pm 0.45}$ & 68.25${\pm 0.65}$ & 57.22${\pm 0.41}$ & 17.15${\pm 0.59}$ & 62.13${\pm 0.35}$ & 60.11${\pm 0.48}$ & 15.42${\pm 0.54}$ & 64.32${\pm 0.51}$ \\
DER++ & 71.45${\pm 0.72}$ & 9.62${\pm 0.45}$ & 75.14${\pm 0.65}$ & 68.02${\pm 0.58}$ & 5.74${\pm 0.32}$ & 67.70${\pm 0.55}$ & 69.81${\pm 0.48}$ & 7.82${\pm 0.35}$ & 72.15${\pm 0.52}$ & 61.43${\pm 0.38}$ & 12.53${\pm 0.44}$ & 66.41${\pm 0.32}$ & 64.44${\pm 0.42}$ & 11.34${\pm 0.48}$ & 68.80${\pm 0.45}$ \\
ER-ACE & 69.73${\pm 0.78}$ & 10.85${\pm 0.49}$ & 73.32${\pm 0.72}$ & 66.44${\pm 0.65}$ & 6.42${\pm 0.38}$ & 66.22${\pm 0.62}$ & 68.21${\pm 0.55}$ & 8.93${\pm 0.42}$ & 70.75${\pm 0.58}$ & 59.83${\pm 0.42}$ & 14.82${\pm 0.52}$ & 64.71${\pm 0.36}$ & 62.74${\pm 0.46}$ & 13.14${\pm 0.52}$ & 67.11${\pm 0.48}$ \\
CSReL-CL & 71.92${\pm 0.82}$ & 9.25${\pm 0.52}$ & 75.54${\pm 0.76}$ & 67.44${\pm 0.69}$ & 6.12${\pm 0.42}$ & 67.23${\pm 0.65}$ & 69.42${\pm 0.59}$ & 8.24${\pm 0.46}$ & 71.83${\pm 0.62}$ & 60.74${\pm 0.45}$ & 13.42${\pm 0.56}$ & 65.51${\pm 0.39}$ & 63.83${\pm 0.49}$ & 12.05${\pm 0.56}$ & 67.94${\pm 0.53}$ \\
\midrule
\multicolumn{16}{l}{\emph{Explanation-aware}} \\
RRR & 72.21${\pm 0.78}$ & 9.10${\pm 0.44}$ & 75.81${\pm 0.73}$ & 68.25${\pm 0.33}$ & 5.66${\pm 1.06}$ & 67.95${\pm 0.54}$ & 70.13${\pm 0.65}$ & 7.66${\pm 0.62}$ & 72.43${\pm 0.50}$ & 61.73${\pm 0.43}$ & 12.06${\pm 0.21}$ & 66.68${\pm 0.31}$ & 64.82${\pm 0.32}$ & 10.87${\pm 0.35}$ & 69.16${\pm 0.64}$ \\
RRR-LRP & \underline{73.25${\pm 0.65}$} & \underline{8.42${\pm 0.38}$} & \underline{76.74${\pm 0.58}$} & \underline{69.30${\pm 0.52}$} & \underline{5.13${\pm 1.28}$} & \underline{68.92${\pm 0.49}$} & \underline{71.15${\pm 0.45}$} & \underline{6.92${\pm 0.92}$} & \underline{73.32${\pm 0.46}$} & \underline{63.01${\pm 0.35}$} & \underline{10.24${\pm 0.69}$} & \underline{68.12${\pm 0.28}$} & \underline{66.20${\pm 0.36}$} & \underline{9.23${\pm 0.42}$} & \underline{70.74${\pm 0.41}$} \\
EPR & 72.86${\pm 0.72}$ & 8.75${\pm 0.41}$ & 76.32${\pm 0.65}$ & 68.82${\pm 0.42}$ & 5.36${\pm 0.82}$ & 68.41${\pm 0.51}$ & 70.68${\pm 0.55}$ & 7.21${\pm 0.84}$ & 72.88${\pm 0.48}$ & 62.45${\pm 0.39}$ & 11.15${\pm 0.52}$ & 67.42${\pm 0.30}$ & 65.54${\pm 1.04}$ & 10.02${\pm 0.38}$ & 70.05${\pm 0.52}$ \\
TASS & 70.24${\pm 0.85}$ & 10.35${\pm 0.56}$ & 74.12${\pm 0.78}$ & 66.91${\pm 0.72}$ & 6.74${\pm 0.45}$ & 66.73${\pm 0.68}$ & 68.92${\pm 0.62}$ & 8.63${\pm 0.49}$ & 71.42${\pm 0.65}$ & 60.13${\pm 0.48}$ & 14.25${\pm 0.62}$ & 65.01${\pm 0.42}$ & 63.01${\pm 0.52}$ & 12.83${\pm 0.58}$ & 67.44${\pm 0.55}$ \\
\midrule
\method  & $\mathbf{79.73{\pm 0.31}}$ & $\mathbf{5.32{\pm 0.63}}$ & $\mathbf{82.24{\pm 0.35}}$ & $\mathbf{73.01{\pm 0.23}}$ & $\mathbf{3.14{\pm 0.55}}$ & $\mathbf{72.51{\pm 0.88}}$ & $\mathbf{74.35{\pm 0.46}}$ & $\mathbf{4.23{\pm 0.29}}$ & $\mathbf{75.84{\pm 0.65}}$ & $\mathbf{67.84{\pm 0.72}}$ & $\mathbf{6.52{\pm 0.46}}$ & $\mathbf{70.83{\pm 0.41}}$ & $\mathbf{71.12{\pm 0.53}}$ & $\mathbf{5.82{\pm 0.25}}$ & $\mathbf{72.23{\pm 0.92}}$ \\
\method{}++ & 83.91${\pm 0.28}$ & 4.81${\pm 0.52}$ & 83.18${\pm 0.31}$ & 74.72${\pm 0.19}$ & 2.84${\pm 0.45}$ & 74.83${\pm 0.72}$ & 75.26${\pm 0.38}$ & 3.82${\pm 0.24}$ & 75.48${\pm 0.58}$ & 69.82${\pm 0.61}$ & 5.89${\pm 0.39}$ & 69.47${\pm 0.35}$ & 72.11${\pm 0.44}$ & 5.26${\pm 0.21}$ & 72.55${\pm 0.81}$ \\

\midrule
\multicolumn{16}{c}{\emph{ViT-B}}\\
\midrule
Joint Training & 86.12${\pm 0.23}$ & - & - & 81.31${\pm 0.29}$ & - & - & 84.59${\pm 0.31}$ & - & - & 75.31${\pm 0.19}$ & - & - & 78.96${\pm 0.24}$ & - & - \\
\midrule
\multicolumn{16}{l}{\emph{Prompt-based}} \\
L2P & 70.82${\pm 1.24}$ & 7.94${\pm 0.65}$ & 74.24${\pm 0.91}$ & 71.85${\pm 0.78}$ & 4.23${\pm 0.42}$ & 72.40${\pm 0.83}$ & 71.49${\pm 0.69}$ & 4.85${\pm 0.56}$ & 71.91${\pm 0.74}$ & 54.94${\pm 0.82}$ & 6.83${\pm 0.44}$ & 56.11${\pm 0.57}$ & 64.53${\pm 0.70}$ & 6.24${\pm 0.55}$ & 66.12${\pm 0.69}$ \\
DualPrompt & 72.43${\pm 0.81}$ & 6.82${\pm 0.54}$ & 75.94${\pm 0.69}$ & 73.22${\pm 0.75}$ & 3.74${\pm 0.39}$ & 73.83${\pm 0.71}$ & 72.12${\pm 0.67}$ & 4.33${\pm 0.43}$ & 72.46${\pm 0.86}$ & 54.05${\pm 0.70}$ & 7.24${\pm 0.42}$ & 55.04${\pm 0.45}$ & 66.12${\pm 0.58}$ & 5.52${\pm 0.44}$ & 66.35${\pm 0.57}$ \\
CODA-Prompt & 74.74${\pm 0.79}$ & 5.42${\pm 0.42}$ & 78.13${\pm 0.67}$ & 74.95${\pm 0.82}$ & 3.12${\pm 0.38}$ & 75.50${\pm 0.78}$ & 72.35${\pm 0.65}$ & 3.82${\pm 0.41}$ & 72.69${\pm 0.64}$ & 54.24${\pm 0.68}$ & 6.92${\pm 0.40}$ & 55.49${\pm 0.53}$ & 67.82${\pm 0.66}$ & 4.83${\pm 0.42}$ & 68.24${\pm 0.55}$ \\
\midrule
\multicolumn{16}{l}{\emph{Adapter-based}} \\
MOS & 75.94${\pm 0.68}$ & 4.82${\pm 0.41}$ & 79.53${\pm 0.55}$ & 76.14${\pm 0.70}$ & 2.12${\pm 0.39}$ & 76.81${\pm 0.66}$ & 76.88${\pm 0.24}$ & 4.15${\pm 0.61}$ & 78.35${\pm 0.53}$ & 61.34${\pm 0.68}$ & 3.52${\pm 0.29}$ & 62.26${\pm 0.35}$ & 71.53${\pm 0.67}$ & 3.54${\pm 0.31}$ & 71.82${\pm 0.56}$ \\
MIN & 75.45${\pm 0.83}$ & 5.23${\pm 0.54}$ & 79.12${\pm 0.69}$ & 75.75${\pm 0.85}$ & 2.74${\pm 0.41}$ & 76.30${\pm 0.81}$ & 76.15${\pm 0.68}$ & 4.31${\pm 0.36}$ & 78.54${\pm 0.66}$ & 61.92${\pm 0.81}$ & 3.14${\pm 0.28}$ & 62.42${\pm 0.57}$ & 72.15${\pm 0.55}$ & 3.62${\pm 0.42}$ & 72.61${\pm 0.69}$ \\
PLAN & 76.12${\pm 0.70}$ & 4.83${\pm 0.42}$ & 79.74${\pm 0.56}$ & 76.03${\pm 0.83}$ & 2.24${\pm 0.28}$ & 76.76${\pm 0.67}$ & 76.62${\pm 0.55}$ & 4.02${\pm 0.31}$ & 78.93${\pm 0.84}$ & 60.25${\pm 0.69}$ & 4.52${\pm 0.32}$ & 61.14${\pm 0.55}$ & 70.82${\pm 0.68}$ & 3.42${\pm 0.30}$ & 71.23${\pm 1.03}$ \\
\midrule
\multicolumn{16}{l}{\emph{Replay-based}} \\
RRR-ViT & 76.94${\pm 1.17}$ & 4.82${\pm 0.61}$ & 78.53${\pm 1.14}$ & 78.32${\pm 1.21}$ & 2.01${\pm 0.49}$ & 78.19${\pm 1.16}$ & 78.83${\pm 0.94}$ & 4.12${\pm 0.50}$ & 80.84${\pm 0.93}$ & 60.52${\pm 1.18}$ & 5.43${\pm 0.23}$ & 63.24${\pm 0.64}$ & 70.19${\pm 1.17}$ & 3.82${\pm 0.51}$ & 70.94${\pm 1.16}$ \\
RRR-ViT-AttnLRP & 78.09${\pm 0.85}$ & 4.17${\pm 0.58}$ & 80.34${\pm 0.88}$ & 78.47${\pm 1.08}$ & 1.92${\pm 0.37}$ & 78.56${\pm 0.90}$ & 79.48${\pm 0.76}$ & 3.35${\pm 1.01}$ & 80.87${\pm 0.66}$ & 63.02${\pm 0.56}$ & 4.24${\pm 0.26}$ & 63.51${\pm 0.59}$ & 71.61${\pm 0.87}$ & 3.33${\pm 0.42}$ & 71.88${\pm 0.89}$ \\
L2P-R & 69.62${\pm 1.25}$ & 6.34${\pm 0.56}$ & 77.25${\pm 0.92}$ & 74.23${\pm 0.29}$ & 3.42${\pm 0.53}$ & 74.82${\pm 1.04}$ & 76.82${\pm 0.90}$ & 3.91${\pm 0.57}$ & 77.34${\pm 0.89}$ & 58.33${\pm 0.23}$ & 6.14${\pm 0.21}$ & 59.71${\pm 0.68}$ & 67.24${\pm 1.21}$ & 5.32${\pm 0.56}$ & 68.15${\pm 0.61}$ \\
CODA-R & 77.53${\pm 0.99}$ & 4.23${\pm 0.42}$ & 80.24${\pm 0.77}$ & 78.02${\pm 0.82}$ & 3.02${\pm 0.28}$ & 78.21${\pm 0.78}$ & 78.54${\pm 0.66}$ & 2.82${\pm 0.31}$ & 79.52${\pm 0.64}$ & 61.24${\pm 0.89}$ & 4.82${\pm 0.42}$ & 61.43${\pm 0.55}$ & 71.53${\pm 0.78}$ & 3.24${\pm 0.30}$ & 71.62${\pm 0.67}$ \\
MOS-R & 78.82${\pm 0.65}$ & 3.84${\pm 0.29}$ & \underline{82.52${\pm 0.53}$} & \underline{79.03${\pm 0.78}$} & \underline{1.52${\pm 0.42}$} & \underline{79.42${\pm 0.54}$} & \underline{80.92${\pm 0.42}$} & \underline{2.43${\pm 0.28}$} & \underline{81.55${\pm 0.31}$} & \underline{67.84${\pm 0.96}$} & \underline{2.24${\pm 0.27}$} & \underline{66.25${\pm 0.42}$} & \underline{73.62${\pm 0.35}$} & \underline{2.52${\pm 0.29}$} & \underline{73.89${\pm 0.93}$} \\
PLAN-R & \underline{79.24${\pm 0.56}$} & \underline{3.52${\pm 0.28}$} & 82.14${\pm 0.54}$ & 78.62${\pm 0.59}$ & 1.83${\pm 0.27}$ & 78.92${\pm 0.65}$ & {80.13${\pm 0.51}$} & 2.58${\pm 1.29}$ & {80.89${\pm 0.42}$} & 65.48${\pm 0.57}$ & 3.05${\pm 0.31}$ & 65.83${\pm 0.51}$ & 72.85${\pm 0.56}$ & 2.84${\pm 0.28}$ & 73.06${\pm 0.55}$ \\
\midrule
\method  & $\mathbf{83.65{\pm 0.28}}$ & $\mathbf{2.92{\pm 0.24}}$ & $\mathbf{84.64{\pm 0.37}}$ & $\mathbf{80.43{\pm 0.55}}$ & $\mathbf{1.23{\pm 0.62}}$ & $\mathbf{80.82{\pm 0.26}}$ & $\mathbf{82.74{\pm 0.76}}$ & $\mathbf{1.89{\pm 0.33}}$ & $\mathbf{83.92{\pm 0.45}}$ & $\mathbf{71.14{\pm 0.59}}$ & $\mathbf{1.53{\pm 0.41}}$ & $\mathbf{70.93{\pm 0.38}}$ & $\mathbf{76.83{\pm 0.56}}$ & $\mathbf{1.92{\pm 0.83}}$ & $\mathbf{77.92{\pm 0.78}}$ \\

\method{}++ & 87.21${\pm 0.22}$ & 2.62${\pm 0.20}$ & 86.79${\pm 0.31}$ & 81.02${\pm 0.48}$ & 1.10${\pm 0.55}$ & 80.89${\pm 0.21}$ & 83.15${\pm 0.67}$ & 1.72${\pm 0.29}$ & 84.21${\pm 0.38}$ & 73.58${\pm 0.51}$ & 1.37${\pm 0.36}$ & 73.81${\pm 0.32}$ & 77.12${\pm 0.49}$ & 1.72${\pm 0.74}$ & 78.48${\pm 0.69}$ \\

\bottomrule
\end{tabular}
}
\end{table*}

\section{Additional Experimental Details}
\label{sec:app_xai_qual}

\subsection{Explanation Quality}

\paragraph{Motivation and Protocol Overview.}

To assess whether \method grounds attributions in semantically meaningful features rather than spurious artifacts, we adopt two complementary evaluation protocols.
(1)~\textbf{Spurious Correlation Mitigation:} We measure the model’s ability to ignore spurious biases using final accuracy on FITZ~\cite{bayasi2024biaspruner} and Split ImageNet-100C~\cite{lee2024continual}, following standard evaluation practices in prior debiasing work~\cite{bayasi2024biaspruner,lee2024continual}.
(2)~\textbf{Semantic Localization:} We evaluate whether explanations remain focused on semantic object regions and avoid explanation drift by computing segmentation-based metrics on saliency maps, following~\citet{chefer2021transformer}, on the CUB dataset, which provides ground-truth segmentation annotations\footnote{\url{https://data.caltech.edu/records/w9d68-gec53}}.

\paragraph{Datasets.}
Split ImageNet-100C (I-100C)~\cite{lee2024continual} is derived from Split ImageNet-100, which consists of 10 sequential tasks, each containing 10 disjoint classes randomly sampled from ImageNet-1K. To introduce controlled spurious correlations, we inject Gaussian noise disproportionately: $(\alpha \times 100)\%$ of the samples from one randomly selected class are corrupted, while only 1\% of samples from the remaining classes are affected, for a skew ratio $\alpha \geq 0.01$. We consider seven bias levels by varying the skew ratio.

Fitzpatrick17k (FITZ)~\cite{groh2021evaluating} is a large-scale clinical dermatology dataset comprising 16,577 images across 114 skin conditions, each annotated with Fitzpatrick Skin Type (FST) on a six-point scale. Due to its pronounced skin-tone imbalance-approximately 47\% of images correspond to FST I–II and only about 13\% to FST V–VI-the dataset is widely used to evaluate robustness and bias mitigation under real-world demographic skew.

\paragraph{Training Configuration.}
We employ the ResNet-18 and ViT-B backbones pre-trained on ImageNet-1K and ImageNet-21k respectively.
For spurious correlation mitigation, we finetune ResNet-18 on both FITZ and I-100C using the exact same setup reported in the original codebase of every baseline. Specifically, for BGS, we used its best performing configuration on the accuracy metric reported in the paper, which is DER+BGS. For semantic localization, both models are fine-tuned on the CUB using the same protocol from the main paper.

\paragraph{Semantic Localization Evaluation.}
Here, our goal is to asses the quality of semantic localization and explanation map consistency.
We evaluate explanation quality using the CUB dataset, which provides ground-truth object segmentation masks. Following~\citet{chefer2021transformer}, we threshold saliency maps at the 50th percentile to produce binary masks and compute:  (1)~\textbf{mIoU}: Mean Intersection-over-Union between explanation masks and ground-truth objects,
    (2)~\textbf{pixAcc}: Pixel-wise accuracy (correctly classified pixels / total pixels), and
    (3)~\textbf{mAP}: Mean Average Precision across IoU thresholds $[0.5, 0.55, \ldots, 0.95]$.
Higher values indicate that explanations better localize semantic object regions rather than spurious backgrounds. Since \method is trained to align relevance with concept regions on the object, improved localization is partly expected by design; the spurious-correlation results on FITZ and I-100C provide complementary evidence that does not rely on segmentation.

\subsection{Robustness Tests Further Details}

\paragraph{OOD Evaluation Protocol.} Our setup is inspired by~\cite{yosinski2014transferable}, which studies feature transferability by training on a subset of ImageNet classes; here, we fine-tune on a class subset and evaluate on OOD benchmarks.

\paragraph{Dataset Construction.} We curate a lightweight dataset by sampling 5 images from 500 randomly selected ImageNet-1K~\cite{deng2009imagenet} classes resulting with 2,500 samples. This ensures broad semantic coverage while rigorously testing data efficiency under limited supervision.

\paragraph{Models and Benchmarks.} We utilize ImageNet-pretrained ResNet-18 and ViT-B models. Testing is performed on ImageNet-A and ObjectNet (restricted to overlapping classes), which represent challenging OOD shifts that ImageNet-pretrained models mostly fail with.
Performance is measured using classification accuracy (\%) and mean Intersection over Union (mIoU). For mIoU, we compare the explanation maps, computed as in Tab.~\ref{tab:segmentation}, against object segmentations generated by prompting SAM~3 with the ground-truth class labels.

\paragraph{Training.} We execute the complete \method framework-comprising concept discovery, validation, and grounding - on the 2,500-sample subset. First, we perform automatic concept generation~\cite{oikarinen2023label} for the 500 selected classes. We then validate these candidate concepts using 40 images per class, yielding a filtered concept set for each category. Subsequently, we ground these concepts for every sample in the training subset by prompting SAM~3 with the filtered concepts (one prompt per concept), thereby generating concept maps. Additionally, we employ LRP and AttnLRP to produce explanations for each image. Finally, the models are fine-tuned exclusively on this subset using the \method approach, leveraging the derived concept maps and explanations as supervision.
For both RRR and RRR-ViT we used the same setup reported by the authors~\cite{ebrahimi2020remembering} for the fine-tuning phase.

\paragraph{Full results.}
We additionally evaluate EPR and TASS, including their ViT-adapted variants (EPR-ViT, TASS-ViT), to enable a comprehensive comparison with explanation-aware baselines. The ViT adaptations follow standard practice, using Grad-CAM for ViTs~\cite{chefer2021transformer} and generating explanations from the \texttt{[CLS]} token, as commonly adopted in prior work.

The results mirror the trends observed in the main paper: \method consistently outperforms all explanation-aware methods in robustness to out-of-distribution data.

\begin{table}[h]
    \centering
    \small
    \caption{Out-of-distribution robustness evaluation. We report accuracy (\%) and mIoU on ImageNet-A and ObjectNet before and after fine-tuning with \method using a sparse subset. We compare EPR, TASS, RRR, and C$^{3}$L (for ViT-B, the ViT variants EPR-ViT, TASS-ViT, and RRR-ViT).}
    \label{tab:robustness_results_full}
    \begin{tabular}{lllc@{\hspace{12pt}}cccc}
        \toprule
        \multirow{2}{*}{Model} & \multirow{2}{*}{Dataset} & \multirow{2}{*}{Metric} & \multirow{2}{*}{Before} & \multicolumn{4}{c}{After} \\
        \cmidrule(lr){5-8}
        & & & & EPR & TASS & RRR & \textbf{C$^{3}$L} \\
        \midrule
        \multirow{4}{*}{ResNet-18}
        & \multirow{2}{*}{ImageNet-A} & Acc & 8.59 & 8.49 & 6.34 & 7.92 & \textbf{20.15} \\
        &                            & mIoU & 21.32 & 21.78 & 20.83 & 20.39 & \textbf{43.27} \\
        \cmidrule(l){2-8}
        & \multirow{2}{*}{ObjectNet}  & Acc & 21.88 & 23.08 & 21.62 & 21.56 & \textbf{36.93} \\
        &                            & mIoU & 35.41 & 35.27 & 34.68 & 34.16 & \textbf{52.91} \\
        \midrule
        \multirow{4}{*}{ViT-B}
        & \multirow{2}{*}{ImageNet-A} & Acc & 13.48 & 15.34 & 12.91 & 13.18 & \textbf{30.82} \\
        &                            & mIoU & 26.38 & 26.53 & 25.82 & 25.43 & \textbf{45.32} \\
        \cmidrule(l){2-8}
        & \multirow{2}{*}{ObjectNet}  & Acc & 33.92 & 34.26 & 32.43 & 34.49 & \textbf{58.04} \\
        &                            & mIoU & 38.12 & 37.65 & 37.19 & 37.02 & \textbf{54.83} \\
        \bottomrule
    \end{tabular}
\end{table}

\section{Foundation Models Usage and Fairness Clarification}
\label{sec:app_foundation}

\subsection{Performance without Foundation Models Usage}
\label{sec:app_foundation_gains}

To disentangle the semantic guidance and foundation models factors, we remove foundation models from the training pipeline and instead conduct experiments on the \textbf{PartImageNet} dataset~\cite{he2022partimagenet}, which provides ground-truth part-level annotations. Although part annotations are not equivalent to semantic concepts, this setting offers a controlled evaluation that isolates the effect of semantic guidance. The results therefore illustrate the benefits of leveraging semantic features for continual learning independently of foundation models.

While we aim to ensure a fair comparison with existing baselines, we note that equipping those methods with the same foundation-model guidance used in our approach would be conceptually similar to applying our method or to the experimental setting presented in Table~\ref{tab:concept_vs_object}.

\paragraph{Experimental Setup.}
We evaluate a ResNet-18 trained from scratch on a class-incremental split of PartImageNet. We used 150 classes (out of 158, removing 8 randomly sampled classes) partitioned into 10 tasks of 15 classes each.\\ \textbf{Baselines:} Standard ER \cite{riemer2018learning}, DER++ \cite{buzzega2020dark}, RRR \cite{ebrahimi2020remembering}, EPR \cite{saha2023saliency}, and our original \method.\\
\textbf{\method-PG (Part-Guided):} Instead of using GPT-4o-mini or SAM 3, we used the dataset's ground-truth part masks as the semantic concepts for our dual regularization objective.

\paragraph{Results and Discussion.}
Table \ref{tab:partimagenet_results} demonstrates that even when trained from scratch without foundation model priors, concept-level guidance outperforms all baselines.

\begin{table}[h]
\centering
\caption{Continual Learning on PartImageNet. Comparison of concept-guided learning against state-of-the-art baselines.}
\label{tab:partimagenet_results}
\small
\begin{tabular}{lccc}
\toprule
{Method} & {ACC} $\uparrow$ & {F} $\downarrow$ & {AIA} $\uparrow$ \\ \midrule
ER & 44.21 $\pm$ 1.12 & 21.45 $\pm$ 0.88 & 46.92 $\pm$ 0.94 \\
DER++ & 49.82 $\pm$ 0.75 & 16.32 $\pm$ 0.54 & 52.14 $\pm$ 0.61 \\
RRR & 51.34 $\pm$ 1.04 & 14.92 $\pm$ 0.91 & 53.88 $\pm$ 1.02 \\
EPR & 52.65 $\pm$ 0.82 & 13.51 $\pm$ 0.44 & 55.07 $\pm$ 0.73 \\ \midrule
\method & 59.12 $\pm$ 0.64 & 10.36 $\pm$ 0.37 & 62.43 $\pm$ 0.92 \\
{\method-PG} & {58.83 $\pm$ 0.95} & {10.78 $\pm$ 0.62} & {62.11 $\pm$ 0.88} \\ \bottomrule
\end{tabular}
\end{table}

The performance gap between \method and pixel-based methods confirms that semantic guidance is a primary driver for stability. By grounding the model in persistent parts, we achieve superior retention and learning dynamics, independent of the foundation model used for discovery.

\subsection{Semantic Depth Over Object Segmentation}
\label{sec:app_segconc}
We examined the impact of mask granularity in Appendix~\ref{sec:app_additional_abl} (Table~\ref{tab:concept_vs_object}) to further emphasize the contribution of our semantic anchoring approach.

We implemented a baseline using the exact \textbf{same SAM 3 foundation model} but prompted it for object-level segmentation (masking the entire object) rather than our fine-grained concept-level segmentation (masking specific attributes like ``beak'' or ``wing''). The results demonstrate that concept-level supervision yields superior performance compared to object-level masks. This suggests that semantic guidance is also a driver of our results, rather than merely the presence of foundation-model-derived segmentation masks.

\subsection{Foundation Models as Automated Annotators}
Our framework utilizes GPT-4o-mini and SAM 3 to automate the generation of semantic supervision. We leverage these foundation models in an offline preprocessing stage to decompose images into semantic regions. We view this process as automated annotation leveraging foundation model priors, distinct from knowledge distillation in that we do not transfer class predictions or latent representations.
Importantly, this automated concept discovery and grounding stage is as an offline pre-processing stage.

In standard knowledge distillation, a student model is typically trained to mimic the soft targets (logits) or align with the latent feature space of a teacher model. In contrast, \method does not access the class probability distributions or internal embeddings of the foundation models. Instead, these models function as an automated framework to generate binary spatial masks ($M_{concept}$). These masks provide the "where" (spatial attention targets), but the model must autonomously learn the "what" (discriminative features) from the training data labels. This allows us to inject human-interpretable priors without leaking instance-specific predictive information.

\section{Qualitative Visualizations}
\label{sec:app_qualitative}

\begin{figure}[t]
  \centering
  \includegraphics[width=0.49\textwidth]{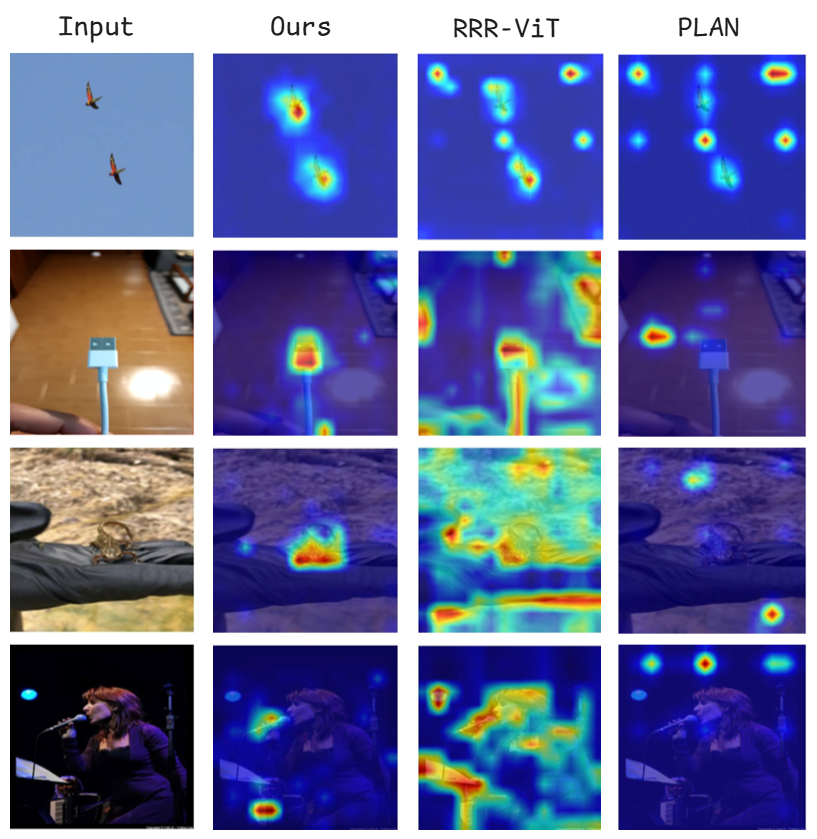}
  \caption{Qualitative comparison of explanation maps produced by \method and baseline methods before and after continual learning. Explanation maps produced using ViT-B w.r.t. the classes (top to bottom): ‘lorikeet’, 'usb cable’, ‘common newt, Triturus vulgaris’, and ‘accordion, piano accordion’.
}
  \label{fig:explanation_viz}
\end{figure}

Figure~\ref{fig:explanation_viz} presents qualitative examples (ImageNet pretrained data) of explanation maps generated by \method and baseline methods before and after continual learning, using a ViT-B model pretrained on ImageNet and trained on the ObjectNet dataset. \method preserves attention on semantically meaningful object regions (e.g., discriminative parts and attributes), whereas baseline methods exhibit explanation drift toward background regions and texture cues.

\section{Failure Cases}
\label{sec:app_failure}

\subsection{Concept Discovery Failures}

While automated concept discovery achieves quite high retention rate, some generated concepts fail filtering:

\paragraph{Too Abstract.} Concepts like ``graceful posture", ``aggressive demeanor", or "natural habitat" describe semantic properties but lack clear visual grounding. SAM 3 cannot reliably localize such concepts, leading to noisy masks.

\paragraph{Too Rare.} Concepts appearing in $<15\%$ of class instances (e.g., ``open beak" for birds, ``raised paw" for dogs) fail occurrence rate criteria. While visually grounded when present, their rarity makes them unsuitable for class-wide regularization.

\paragraph{Ambiguous.} Concepts like ``medium size" or ``dark color" lack clear spatial boundaries. For instance, SAM 3 may  produce inconsistent masks for "dark color" depending on lighting.

\subsection{Segmentation Failures}

SAM 3 rarely produces noisy or incorrect masks:

\paragraph{Small or Occluded Features.} Fine-grained concepts covering $<2\%$ of image area (e.g., ``eye-ring" on small birds, ``toe claws" on animals) are missed or segmented with low IoU. This primarily affects fine-grained datasets like CUB.

\paragraph{Concept Ambiguity.} Polysemous concepts (e.g., ``fluffy texture" can describe fur, clouds, or cotton) may segment unintended regions. While validation filters some cases, occasional false positives remain.

\section{Limitations and Future Work}
\label{sec:app_limitations}

\textbf{Limitations.} While \method advances continual learning through semantic grounding, several limitations warrant consideration. First, our approach relies on foundation models for automated concept generation. Although such models are increasingly used to guide learning in related domains~\cite{prasse2024dcbm,rao2024discover,benou2025show} and have driven recent breakthroughs in many, they may also introduce failure modes. For instance, spatial grounding may fail if language models produce non-visual or ambiguous concepts. We mitigate this through occurrence rate and coverage validation, but more sophisticated concept quality metrics could improve robustness. Second, \method adds computational overhead for the concept and explanation map preprocessing procedure. Yet, this cost is modest relative to the performance gains. Third, our evaluation focuses on vision tasks. Extending concept-based regularization to other modalities (language, audio, multimodal) requires domain-specific concept discovery mechanisms. Fourth, the reference saliency used by $\mathcal{L}_{\text{stab}}$ is model-derived: while its magnitude is free to decay under crowding-out, the relative spatial pattern of background relevance at buffer insertion is preserved, so spurious background cues already present at that point are attenuated rather than fully removed.

\textbf{Broader impact.} Beyond technical contributions, our work has important implications for deploying continual learning systems in safety-critical domains. By maintaining interpretable reasoning patterns alongside predictive accuracy, \method enables practitioners to monitor whether models learn reliably or develop hidden vulnerabilities through spurious correlations. This transparency is crucial for applications like medical diagnosis, autonomous systems, and content moderation where understanding \emph{how} decisions are made is as important as ensuring they are correct. Moreover, our automated framework demonstrates that semantic grounding need not require extensive human supervision, making interpretable continual learning more accessible to practitioners without domain expertise in explainable AI.

\textbf{Future directions.}
This work opens several promising avenues for future research. First, extending concept-level regularization to domain-incremental and task-free (online) continual learning settings may better address scenarios with gradually shifting input distributions or unknown task boundaries. Second, exploring concept compositionality, namely, whether concepts learned for one class can be reused or recombined for related classes in subsequent tasks, could enable more efficient and scalable knowledge accumulation. Third, integrating concept-based regularization with parameter-efficient adaptation techniques for large vision--language models may combine the benefits of semantic grounding with efficient fine-tuning. Fourth, developing theoretical frameworks that characterize when and why concept-level regularization outperforms pixel-level approaches would provide principled guidance for the design of explanation-aware continual learning methods. Additionally, leveraging ``around'' prompts~\citep{oikarinen2023label} to selectively preserve informative background regions, rather than suppressing all background content, may further reduce reliance on spurious correlations. Investigating whether concept-level regularization can support the \emph{detection} of distribution shifts, by monitoring changes in concept activation patterns, could enable adaptive systems that identify novel contexts and trigger human intervention when needed. Finally, we believe that extending this approach to knowledge distillation presents a promising direction; in scenarios where a teacher model relies heavily on spurious correlations, concept-guided regularization could actively intervene during distillation, allowing the student to extract valuable predictive knowledge while strictly suppressing the inheritance of the teacher's flawed reasoning patterns.

\newpage
\section*{NeurIPS Paper Checklist}

\begin{enumerate}

\item {\bf Claims}
    \item[] Question: Do the main claims made in the abstract and introduction accurately reflect the paper's contributions and scope?
    \item[] Answer: \answerYes{}
    \item[] Justification: The abstract and introduction clearly outline our proposed method \method and the theoretical and empirical claims we make regarding the mitigation of explanation drift in continual learning.

\item {\bf Limitations}
    \item[] Question: Does the paper discuss the limitations of the work performed by the authors?
    \item[] Answer: \answerYes{}
    \item[] Justification: We fully discuss limitations of the work and failure cases on Appendices~\ref{sec:app_limitations} and~\ref{sec:app_failure}.

\item {\bf Theory assumptions and proofs}
    \item[] Question: For each theoretical result, does the paper provide the full set of assumptions and a complete (and correct) proof?
    \item[] Answer: \answerNA{}
    \item[] Justification: Our paper focuses on an empirical methodology and does not introduce formal mathematical theorems or proofs; however, we do formulate an approximate relevance conservation constraint in Section 3.3.

    \item {\bf Experimental result reproducibility}
    \item[] Question: Does the paper fully disclose all the information needed to reproduce the main experimental results of the paper to the extent that it affects the main claims and/or conclusions of the paper (regardless of whether the code and data are provided or not)?
    \item[] Answer: \answerYes{}
    \item[] Justification: Section~\ref{sec:app_implementation} details all reproducibility information needed. We also provided the full code in the supplementary.

\item {\bf Open access to data and code}
    \item[] Question: Does the paper provide open access to the data and code, with sufficient instructions to faithfully reproduce the main experimental results, as described in supplemental material?
    \item[] Answer: \answerYes{}
    \item[] Justification: The code is fully readable and we provide a Readme.md file in it in which we explain how to use it.

\item {\bf Experimental setting/details}
    \item[] Question: Does the paper specify all the training and test details (e.g., data splits, hyperparameters, how they were chosen, type of optimizer) necessary to understand the results?
    \item[] Answer: \answerYes{}
    \item[] Justification: Detailed experimental settings, including datasets, baselines, hyperparameters, and training protocols, are provided in Section 4.1 and comprehensively expanded in Appendix~\ref{sec:app_implementation}.

\item {\bf Experiment statistical significance}
    \item[] Question: Does the paper report error bars suitably and correctly defined or other appropriate information about the statistical significance of the experiments?
    \item[] Answer: \answerYes{}
    \item[] Justification: We report the mean and standard deviation across 5 independent runs for our experiments and assess statistical significance using paired t-tests, as stated in Section 4.1.

\item {\bf Experiments compute resources}
    \item[] Question: For each experiment, does the paper provide sufficient information on the computer resources (type of compute workers, memory, time of execution) needed to reproduce the experiments?
    \item[] Answer: \answerYes{}
    \item[] Justification: We detail the compute hardware (NVIDIA A100 GPUs, single-GPU training) in Appendix~\ref{sec:app_implementation}.

\item {\bf Code of ethics}
    \item[] Question: Does the research conducted in the paper conform, in every respect, with the NeurIPS Code of Ethics \url{https://neurips.cc/public/EthicsGuidelines}?
    \item[] Answer: \answerYes{}
    \item[] Justification: We have reviewed the NeurIPS Code of Ethics and confirm that our research and methodology conform to these guidelines.

\item {\bf Broader impacts}
    \item[] Question: Does the paper discuss both potential positive societal impacts and negative societal impacts of the work performed?
    \item[] Answer: \answerYes{}
    \item[] Justification: We discuss our work's broader impact on Appendix~\ref{sec:app_limitations}.

\item {\bf Safeguards}
    \item[] Question: Does the paper describe safeguards that have been put in place for responsible release of data or models that have a high risk for misuse (e.g., pre-trained language models, image generators, or scraped datasets)?
    \item[] Answer: \answerNA{}
    \item[] Justification: Our work deals with continual learning models for computer vision, thus we believe it does not pose such risk.

\item {\bf Licenses for existing assets}
    \item[] Question: Are the creators or original owners of assets (e.g., code, data, models), used in the paper, properly credited and are the license and terms of use explicitly mentioned and properly respected?
    \item[] Answer: \answerYes{}
    \item[] Justification: All existing datasets, foundation models, and baselines used in our work are properly credited and referenced in the main text and appendix.

\item {\bf New assets}
    \item[] Question: Are new assets introduced in the paper well documented and is the documentation provided alongside the assets?
    \item[] Answer: \answerYes{}
    \item[] Justification: Our code is released and is well documented.

\item {\bf Crowdsourcing and research with human subjects}
    \item[] Question: For crowdsourcing experiments and research with human subjects, does the paper include the full text of instructions given to participants and screenshots, if applicable, as well as details about compensation (if any)?
    \item[] Answer: \answerNA{}
    \item[] Justification: The paper does not involve crowdsourcing or research with human subjects.

\item {\bf Institutional review board (IRB) approvals or equivalent for research with human subjects}
    \item[] Question: Does the paper describe potential risks incurred by study participants, whether such risks were disclosed to the subjects, and whether Institutional Review Board (IRB) approvals (or an equivalent approval/review based on the requirements of your country or institution) were obtained?
    \item[] Answer: \answerNA{}%
    \item[] Justification: The paper does not involve crowdsourcing or research with human subjects.


\item {\bf Declaration of LLM usage}
    \item[] Question: Does the paper describe the usage of LLMs if it is an important, original, or non-standard component of the core methods in this research? Note that if the LLM is used only for writing, editing, or formatting purposes and does \emph{not} impact the core methodology, scientific rigor, or originality of the research, declaration is not required.
    \item[] Answer: \answerYes{}
    \item[] Justification: We describe our use of an LLM (GPT-4o-mini) for automated concept discovery as a core component of our methodology in Section 3.2 and Appendix~\ref{sec:app_implementation}.

\end{enumerate}

\end{document}